%% file: main.tex
\documentclass{article}
\pdfoutput=1 

\input{"./preamble/packages"}
\input{"./preamble/definition"}

\input{"./preamble/acronym"}

\usepackage[nonatbib, final]{neurips_2020}

\newcommand{\tbw}[1]{{\color{red} (\textbf{Tania:} #1)}}
\newcommand{\kpm}[1]{{\color{BurntOrange} (\textbf{Kevin:} #1)}}

\newcommand{\balaji}[1]{{\color{blue} (\textbf{Balaji:} #1)}}
\newcommand{\dpk}[1]{{\color{OliveGreen} (\textbf{Deepak:} #1)}}

\newcommand{\shr}[1]{{\color{BurntOrange} (\textbf{Shreyas:} #1)}}
\newcommand{\lzi}[1]{{\color{pink} (\textbf{Zi:} #1)}}

\newcommand{\js}[1]{{\color{teal} (\textbf{Jasper:} #1)}}
\newcommand{\dt}[1]{{\color{purple} (\textbf{Dustin:} #1)}}
\newcommand{\tn}[1]{{\color{Fuchsia} (\textbf{Timothy:} #1)}}

\newcommand{\softmax}{\mathsf{softmax}}
\renewcommand{\logit}{\mathsf{logit}}

\usepackage{ragged2e}

\DeclareTextFontCommand{\emph}{\em}

  \usepackage[compact]{titlesec}
  \titlespacing{\section}{0pt}{1ex}{0ex}
  \titlespacing{\subsection}{0pt}{1ex}{0ex}
  \titlespacing{\subsubsection}{0pt}{0.5ex}{0ex}

\title{
Simple and Principled Uncertainty Estimation with Deterministic Deep Learning via Distance Awareness}

\author{
Jeremiah Zhe Liu\thanks{Work done at Google Research.}\\
Google Research \& Harvard University \\
{\tt jereliu@google.com} \\
\And
Zi Lin$^\dagger$  \\
Google Research \\
{\tt lzi@google.com}
\And
Shreyas Padhy\thanks{Work done as an Google AI Resident.}  \\
Google Research \\
{\tt shreyaspadhy@google.com}
\And
Dustin Tran\\
Google Research \\
{\tt trandustin@google.com}
\And
Tania Bedrax-Weiss  \\
Google Research \\
{\tt tbedrax@google.com}
\And
Balaji Lakshminarayanan  \\
Google Research \\
{\tt balajiln@google.com}
}

\begin{document}
\etocdepthtag.toc{mtchapter}
\etocsettagdepth{mtchapter}{subsection}
\etocsettagdepth{mtappendix}{none}

\maketitle
\input{"./section/abstract"}

\section{Introduction}
\label{sec:intro}

\input{"./section/intro"}

\section{Distance Awareness: An Important Condition for High-Quality Uncertainty Estimation}
\label{sec:theory}
\input{"./section/theory"}

\section{SNGP: A Simple Approach to Distance-aware Deep Learning}
\label{sec:method}
\input{"./section/method"}

\section{Related Work}
\label{sec:related}
\input{"./section/related"}

\section{Experiments}
\label{sec:exp}
\input{section/experiment}

\section{Conclusion 
}
\label{sec:sum}
\input{section/conclusion}

\clearpage
\input{"./section/acknowledge"}

\section*{Broader Impact}
\input{"./section/impact"}

\bibliographystyle{abbrv}

\clearpage
\appendix
\begin{center}{\textbf{\Large{Supplementary Material:\\
Simple and Principled Uncertainty Estimation with \\ Deterministic Deep Learning via Distance Awareness}}}\end{center}
\etocdepthtag.toc{mtappendix}
\etocsettagdepth{mtchapter}{none}
\etocsettagdepth{mtappendix}{subsection}
{
  \hypersetup{linkcolor=black}
  \tableofcontents
}
\newpage

\section{Method Summary}
\label{sec:method_sum}

\paragraph{Architecture} Given a deep learning model  $\logit(\bx)=g \circ h(\bx)$ with $L-1$ hidden layers of size $\{D_l\}_{l=1}^L$, \gls{SNGP} makes two changes to the model: 
\begin{enumerate}
    \item adding spectral normalization to the hidden weights $\{\bW_l\}_{l=1}^L$, and
    \item replacing the dense output layer $g(h) = h^\top\bbeta$ with a \gls{GP} layer. Under the \gls{RFF} approximation, the \gls{GP} layer is simply a one-layer network with $D_L$ hidden units $g(h) \propto \cos(-\bW_L h_i + \bb_L)^\top\bbeta$, where $\{\bW_L, \bb_L\}$ are frozen weights that are initialized from a Gaussian and a uniform distribution, respectively (as described in Equation (\ref{eq:rff_lr})). 
\end{enumerate} 

\paragraph{Training} Algorithm \ref{alg:training} summarizes the training step. As shown, for every minibatch step, the model first updates the hidden-layer weights $\{\bW_l, \bb_l\}_{l=1}^{L-1}$ and the trainable output  weights $\beta=\{\beta_k\}_{k=1}^K$ via \gls{SGD}, 
then performs spectral normalization using power iteration method (\citep{gouk_regularisation_2018} which has time complexity $O(\sum_{l=1}^{L-1} D_l)$), and finally performs precision matrix update (Equation (\ref{eq:gp_posterior}), time complexity $O(D^2_L)$). Since $\{D_l\}_{l=1}^{L-1}$ are fixed for a given architecture and usually $D_L \leq 1024$, the computation scales linearly with respect to the sample size. We use $D_L=1024$ in the experiments.

\paragraph{Prediction} Algorithm \ref{alg:prediction} summarizes the prediction step. The model first performs the conventional forward pass, which involves (i) computing the final hidden feature $\Phi(\bx)_{D_L \times 1}$, and (ii)  computing the posterior mean  $\hat{m}_k(\bx)=\Phi^\top\bbeta_k$ (time complexity $O(D_L)$). Then, using the posterior variance matrices $\{\hat{\bSigma}_k\}_{k=1}^K$, the model computes the predictive variance $\hat{\sigma}_k(\bx)^2=\Phi(\bx)^\top\hat{\Sigma}\Phi(\bx)$ (time complexity $O(D^2_L)$). 

To estimate the predictive distribution $p_k = \exp(m_k) / \sum_k \exp(m_k)$ where $m_k \sim N \big(\hat{m}_k(\bx), \hat{\sigma}_k^2(\bx)\big)$, we calculate its posterior mean using Monte Carlo averaging. Notice that this Monte Carlo averaging is computationally very cheap since it only involves sampling from a closed-form distribution whose parameters $(\hat{m}, \hat{\sigma}^2)$ are already computed by the single feed-forward pass (i.e., a single call to {\small\texttt{tf.random.normal}}). This is different from the full Monte Carlo sampling used by \gls{MCD} or deep ensembles which require multiple forward passes and are computationally expensive. As shown in the latency results in experiments (Section \ref{sec:exp_ml}), the extra variance-related computation only adds a small overhead to the inference time of a deterministic \gls{DNN}. In the experiments, we use 10 samples to compute the mean predictive distribution.

In applications where the inference latency is of high priority (e.g., real-time pCTR prediction for online advertising), we can reduce the computational overhead further by replacing the Monte Carlo averaging with the mean-field approximation \citep{daunizeau_semi-analytical_2017}. We leave this for future work. 


\newpage
\section{Formal Statements}
\label{sec:formal}

\paragraph{Minimax Solution to Uncertainty Risk Minimization}
\label{sec:minimax_app}

The expression in (\ref{eq:minimax_solution}) seeks to answer the following question: \textit{assuming we know the true domain probability $p^*(\bx \in \Xsc_{\texttt{IND}})$, and given a model $p(y|\bx, \bx\in \Xsc_{\texttt{IND}})$ that we have already learned from data, what is the best solution we can construct to minimize the minimax objective (\ref{eq:minimax_loss})?} The interest of this conclusion is not to construct a practical algorithm, but to highlight the theoretical necessity of taking into account the domain probability in constructing a good solution for uncertainty quantification. If the domain probability is not necessary, then the expression of the unique and optimal solution to the minimax probability should not contain $p^*(\bx \in \Xsc_{\texttt{IND}})$ even if it is available. However the expression of (\ref{eq:minimax_solution}) shows this is not the case.

To make the presentation clear, we formalize the statement about (\ref{eq:minimax_solution}) into the below  proposition:
\begin{proposition}[Minimax Solution to Uncertainty Risk Minimization] $ $\\
Given:
\begin{enumerate}
    \item[(a)] $p(y|\bx, \bx \in \Xsc_{\textup{\texttt{IND}}})$ the model's predictive distribution learned from data $\Dsc=\{y_i, \bx_i\}_{i=1}^N$
    \item[(b)] $p^*(\bx \in \Xsc_{\textup{\texttt{IND}}})$ the true domain probability,
\end{enumerate}
then there exists an unique optimal solution to the minimax problem (\ref{eq:minimax_loss}), and it can be constructed using (a) and (b) as:
\begin{align}
    p(y |\bx) 
    &= 
    p(y|\bx, \bx \in \Xsc_{\textup{\texttt{IND}}}) * p^*(\bx \in \Xsc_{\textup{\texttt{IND}}}) 
    + 
    p_{\textup{\texttt{uniform}}}(y|\bx, \bx \not\in \Xsc_{\textup{\texttt{IND}}}) * p^*(\bx \not\in \Xsc_{\textup{\texttt{IND}}})
    \label{eq:minimax_solution_app}
\end{align}
where $p_{\textup{\texttt{uniform}}}(y|\bx, \bx \not\in \Xsc_{\textup{\texttt{IND}}})=\frac{1}{K}$ is a discrete uniform distribution for $K$ classes.
\label{thm:minimax}
\end{proposition}

As discussed in Section \ref{sec:minimax}, the solution (\ref{eq:minimax_solution_app}) is not only optimal for the minimax Brier risk, but is in fact optimal for a wide family of strictly proper scoring rules known as the (separable) \textit{Bregman score} \citep{parry_proper_2012}:
\begin{align}
    s(p, p^*|\bx)= \sum_{k=1}^K 
    \Big\{
    [p^*(y_k|\bx) - p(y_k|\bx)]\psi'(p^*(y_k|\bx)) - \psi(p^*(y_k|\bx))
    \Big\}
    \label{eq:bregman_score}
\end{align}
where $\psi$ is a strictly concave and differentiable function. Bregman score reduces  to the log score when $\psi(p)=p * \log (p)$, and reduces to the Brier score when $\psi(p)=p^2 - \frac{1}{K}$. 

Therefore we will show (\ref{eq:minimax_solution_app}) for the Bregman score. The proof relies on the following key lemma:

\begin{lemma}[$p_{\texttt{uniform}}$ is Optimal for Minimax Bregman Score in $\bx \not\in \Xsc_{IND}$] $ $\\
Consider the Bregman score in (\ref{eq:bregman_score}). At a location $\bx \not\in \Xsc_{IND}$ where the model has no information about $p^*$ other than $\sum_{k=1}^K p(y_k|\bx) = 1$, the solution to the minimax problem
$$\inf_{p\in \Psc} \sup_{p^* \in \Psc^*} s(p, p^*|\bx)$$
is the discrete uniform distribution, i.e., $p_{\texttt{uniform}}(y_k|\bx)=\frac{1}{K} \forall k \in \{1, \dots, K\}$.
\label{thm:minimax_lemma}
\end{lemma}
The proof for Lemma \ref{thm:minimax_lemma} is in Section \ref{sec:minimax_lemma_proof}. It is worth noting that Lemma \ref{thm:minimax_lemma} only holds for a \textit{strictly} proper scoring rule \citep{gneiting_probabilistic_2007}. For a non-strict proper scoring rule (e.g., the \gls{ECE}), there can exist infinitely many optimal solutions, making the minimax  problem ill-posed. 

We are now ready to prove Proposition \ref{thm:minimax}:
\begin{proof}
Denote $\Xsc_{\texttt{OOD}} = \Xsc / \Xsc_{\texttt{IND}}$.
Decompose the overall Bregman risk by domain:
\begin{align*}
    S(p, p^*) 
    &= 
    E_{x \in \Xsc}\big(s(p, p^*|\bx )\big) = 
    \int_{\Xsc} s\big(p, p^*|\bx \big)p^*(\bx) d\bx \\
    & = 
    \int_{\Xsc} s\big(p, p^*|\bx \big) * \big[
    p^*(\bx|\bx \in \Xsc_{\texttt{IND}})p^*(\bx \in \Xsc_{\texttt{IND}}) + 
    p^*(\bx|\bx \in \Xsc_{\texttt{OOD}})p^*(\bx \in \Xsc_{\texttt{OOD}}) \big]
    d\bx \\
    &= E_{\bx \in \Xsc_{\texttt{IND}}}\big(s(p, p^*|\bx )\big)
    p^*(\bx \in \Xsc_{\texttt{IND}}) +
    E_{\bx \in \Xsc_{\texttt{OOD}}}\big(s(p, p^*|\bx )\big)
    p^*(\bx \in \Xsc_{\texttt{OOD}}) \\
    &= S_{\texttt{IND}}(p, p^*) * p^*(\bx \in \Xsc_{\texttt{IND}}) +
    S_{\texttt{OOD}}(p, p^*) * p^*(\bx \in \Xsc_{\texttt{OOD}}).
\end{align*}
where we have denoted $S_{\texttt{IND}}(p, p^*)=E_{\bx \in \Xsc_{\texttt{IND}}}\big(s(p, p^*|\bx )\big)$ and $S_{\texttt{OOD}}(p, p^*)=E_{\bx \in \Xsc_{\texttt{OOD}}}\big(s(p, p^*|\bx )\big)$.

Now consider decomposing the sup risk $\sup_{p^*} S(p, p^*)$ for a given $p$. Notice that sup risk $\sup_{p^*} S(p, p^*)$ is separable by domain for any $p \in \Psc$. This is because $S_{\texttt{IND}}(p, p^*)$ and $S_{\texttt{OOD}}(p, p^*)$ has disjoint support, and we do not impose assumption on $p^*$:
\begin{align*}
    \sup_{p^*} S(p, p^*) = 
    \sup_{p^*} \big[S_{\texttt{IND}}(p, p^*)\big] * 
    p^*(\bx \in \Xsc_{\texttt{IND}}) +
    \sup_{p^*} \big[S_{\texttt{OOD}}(p, p^*)\big] * 
    p^*(\bx \in \Xsc_{\texttt{OOD}})
\end{align*}
We are now ready to decompose the minimax risk $\inf_{p}\sup_{p^*} S(p, p^*)$. Notice that the minimax risk is also separable by domain due to the disjoint in  support:
\begin{align}
    \inf_{p}\sup_{p^*} S(p, p^*) 
    &= 
    \inf_{p}\Big[ 
    \sup_{p^*} \big[S_{\texttt{IND}}(p, p^*)\big] * 
    p^*(\bx \in \Xsc_{\texttt{IND}}) +
    \sup_{p^*} \big[S_{\texttt{OOD}}(p, p^*)\big] * 
    p^*(\bx \in \Xsc_{\texttt{OOD}})
    \Big] \nonumber \\
    &= \inf_{p}\sup_{p^*} \big[S_{\texttt{IND}}(p, p^*)\big] * 
    p^*(\bx \in \Xsc_{\texttt{IND}}) + 
    \inf_{p}\sup_{p^*} \big[S_{\texttt{OOD}}(p, p^*)\big] * 
    p^*(\bx \in \Xsc_{\texttt{OOD}}),
    \label{eq:minimax_decomposed}
\end{align}
also notice that the in-domain minimax risk $\inf_{p}\sup_{p^*} \big[S_{\texttt{IND}}(p, p^*)\big]$ is fixed due to condition $(a)$. \\

Therefore, to show that (\ref{eq:minimax_solution_app}) is the optimal and unique solution to (\ref{eq:minimax_decomposed}), we only need to show $p_{uniform}$ is the optimal and unique solution to $\inf_{p}\sup_{p^*} \big[S_{\texttt{OOD}}(p, p^*)\big]$. To this end, notice that for a given $p$:
\begin{align}
    \sup_{p^*\in \Psc^*} \big[S_{\texttt{OOD}}(p, p^*)\big] 
    = 
    \int_{\Xsc_{\texttt{OOD}}} 
    \sup_{p^*}[s(p, p^*|\bx)] p(\bx|\bx\in \Xsc_{\texttt{OOD}}) d\bx,
\end{align}
due to the fact that we don't impose assumption on $p^*$ (therefore $p^*$ is free to attain the global supreme by maximizing $s(p, p^*|\bx)$ at every single location $\bx \in \Xsc_{\texttt{OOD}}$). Furthermore, there exists $p$ that minimize $\sup_{p^*} s(p, p^*|\bx)$ at every location of $\bx\in \Xsc_{\texttt{OOD}}$, then it minimizes the integral \citep{berger_statistical_1985}. By Lemma \ref{thm:minimax_lemma}, such $p$ exists and is unique, i.e.:
\begin{align*}
    p_{uniform} = \underset{p\in \Psc}{\mathrm{arginf}}\sup_{p^*\in \Psc^*} S_{\texttt{OOD}}(p, p^*).
\end{align*}
In conclusion, we have shown that $p_{uniform}$ is the unique solution to $\inf_{p}\sup_{p^*} S_{\texttt{OOD}}(p, p^*)$. Combining with condition (a)-(b), we have shown that the unique solution to (\ref{eq:minimax_decomposed}) is 
(\ref{eq:minimax_solution_app}).
\end{proof}

\section{Experiment Details and Further Results}
\label{sec:exp_app}

\subsection{2D Synthetic Benchmark} 
For both benchmarks, we sample 500 observations $\bx_i=(x_{1i}, x_{2i})$ from each of the two in-domain classes (orange and blue), and consider a deep architecture ResFFN-12-128, which contains 12 residual feedforward layers with 128 hidden units and dropout rate 0.01. The input dimension is projected from 2 dimensions to the 128 dimensions using a dense layer.

In addition to \gls{SNGP}, we also visualize the uncertainty surface of the below approaches: \textbf{\glsfirst{GP}} is a standard Gaussian process directly taking $\bx_i$ as in input. In low-dimensional datasets, \gls{GP} is often considered the gold standard for uncertainty quantification. \textbf{Deep Ensemble} is an ensemble of 10 ResFFN-12-128 models with dense output layers, \textbf{\gls{MCD}} uses single ResFFN-12-128 model with dense output layer and 10 dropout samples. \textbf{DNN-GP} uses a single ResFFN-12-128 model with the \gls{GP} Layer (described in Section \ref{sec:gp}) without spectral normalization. Finally, \textbf{\gls{SNGP}} uses a single ResFFN-12-128 model with the \gls{GP} layer and with the spectral normalization.

For these two binary classification tasks, in Figure \ref{fig:2d_exp} we plot the predictive uncertainty for \textbf{\gls{GP}}, \textbf{DNN-GP} and \textbf{\gls{SNGP}} as the posterior predictive variance of the logits, i.e., $u(\bx) = var(logit(\bx))$ which ranges between $[0, 1]$ under the \gls{RBF} kernel. For \textbf{\gls{MCD}} and \textbf{Deep Ensemble}, since these two methods don't provide an convenient expression of predictive variance, we plot their predictive uncertainty as the distance of the maximum predictive probability from $0.5$, i.e., $u(\bx) = 1 - 2 * |p(\bx) - 0.5|$, so that $u(\bx) \in [0, 1]$. 

Figure \ref{fig:2d_exp_oval_app}-\ref{fig:2d_exp_moon_app} compares the aforementioned methods in terms of the same metric based on the predictive probability introduced in the last paragraph: $u(\bx) = 1 - 2 * |p - 0.5|$. We also included \textbf{DNN-SN} (a ResFFN-12-128 model with spectral normalization but no \gls{GP} layer) into comparison. As shown, compared to the uncertainty surface based on predictive variance (Figure \ref{fig:2d_exp}), the uncertainty surface based on predictive probability shows stronger influence from the model's decision boundary. This empirical observation seems to suggest that the predictive uncertainty from the \gls{GP} logits can be a better metric for calibration and \gls{OOD} detection. We will explore the performance difference of different uncertainty metrics in calibration and \gls{OOD} performance in the future work. 

\input{section/figures/2d_exp_oval_app}
\input{section/figures/2d_exp_moon_app}

\clearpage
\subsection{Vision and Language Understanding}

\paragraph{Hyperparameter Configuration}
\gls{SNGP} is composed of two components: Spectral Normalization (SN) and Gaussian Process (GP) layer, both are available at the open-source Edward2 probabilistic programming library \footnote{https://github.com/google/edward2}.  

Spectral normalization contains two hyperparameters: the number of power iterations and the upper bound for spectral norm (i.e., $c$ in Equation (\ref{eq:spec_norm})). In our experiments, we find it is sufficient to fix power iteration to 1. The value for the spectral norm bound $c$ controls the trade-off between the expressiveness and the distance-awareness of the residual block, where a small value of $c$ may shrink the residual block back to identity mapping hence harming the expressiveness, while a large value of $c$ may lead to the loss of bi-Lipschitz property (Proposition \ref{thm:resnet_lipschitz}). Furthermore, the proper range of $c$ depends on the layer type: for dense layers (e.g., the intermediate and the output dense layers of a Transformer), it is sufficient to set $c$ to a value between $(0.95, 1)$. For the convolutional layers, the norm bound needs to be set to a larger value to not impact the model's predictive performance. This is likely caused by the fact that the current spectral normalization technique does not have a precise control of the true spectral norm of the convolutional kernel, in conjuction with the fact that the other regularization mechanisms (e.g., BatchNorm and Dropout) may rescale a layer's spectral norm in unexpected ways \citep{gouk_regularisation_2018, miyato_spectral_2018}. In general, we recommend performing a grid search for $c \in \{0.9, 1, 2, ...\}$ to identify the smallest possible values of $c$ that still retains the predictive performance of the original model. In the experiments, we set the norm bound to $c=6$ for a WideResNet model.

The \gls{GP} layer contains 3 hyperparameters for the main layer (Equation (\ref{eq:rff_lr})), and 2 hyperparameters for its covariance module (Equation (\ref{eq:gp_posterior}). The three hyperparameter for the main layers are the \textit{hidden dimension} ($D_{L}$, i.e., the number of random features), the \textit{length-scale parameter} $l$ for the \gls{RBF} kernel, and the strength of $L_2$ regularization on output weights $\beta_k$. In the experiments, we find the model's performance to be not very sensitive to these parameters. Setting $D_L=1024$ or $2048$, $l=2.0$ and $L_2$ regularization to 0 are sufficient in most cases. The two hyperparameters for the covariance module is the ridge factor $s$ and the discount factor $m$, they come into the update rule of the precision matrix as:
$$\Sigma_{k, 0}^{-1} = s*\bI, 
\quad 
\Sigma_{k, t}^{-1} = m * \Sigma_{k, t-1}^{-1} + (1-m) * \sum_{i=1}^M \hat{p}_{ik}(1-\hat{p}_{ik})\Phi_i\Phi_i^\top,$$
i.e., the ridge factor $s$ serves to control the stability of matrix inverse (if the number of sample size $n$ is small), and $m$ controls how fast the moving average update converges to the population value $\Sigma_{k}=s\bI + \sum_{i=1}^n \hat{p}_{ik}(1-\hat{p}_{ik})\Phi_i\Phi_i^\top$. Similar to other moving-average update method, these two parameters can impact the quality of learned covariance matrix in non-trivial ways. In general, we recommend conducting some small scale experiments on the data to validate the learning quality of the moving average update in approximating the population covariance. Alternatively, the covariance update can be computed exactly at the final epoch by initialize $\Sigma_{k, 0}^{-1} =\bzero$ and simply using the update formula $\Sigma_{k, t}^{-1} = \Sigma_{k, t-1}^{-1} + \sum_{i=1}^M \hat{p}_{ik}(1-\hat{p}_{ik})\Phi_i\Phi_i^\top$. In the experiments, we set $s=0.001$ and $m=0.999$, which is sufficient for our setting where the number of minibatch steps per epoch is large.

We also implemented two additional functionalities for GP layers:  \textit{input dimension projection} and \textit{input layer normalization}. The input dimension project serves to project the hidden dimension of the penultimate layer $D_{L-1}$ to a lower value $D'_{L-1}$ (using a random Gaussian matrix $\bW_{D_{L-1} \times D'_{L-1}}$), it can be projected down to a smaller dimension. \textit{Input layer normalization} applies Layer Normalization to the input hidden features, which is akin to performing \gls{ARD}-style variable selection to the input features. In the experiments, we always turn on the \textit{input layer normalization} and set input layer normalization to 128. Although later ablation studies revealed that the model performance is not sensitive to these values.

\begin{table}[ht]
\centering
\scalebox{0.65}{
\begin{tabular}{|c|c|c|c|}
\toprule
 \multicolumn{2}{c}{Spectral Normalization}  &  
 \multicolumn{2}{c}{Gaussian Process Layer}  \\ \midrule
Power Iteration & 1  &  
Hidden Dimension & 1024 (WRN), 2048 (BERT)  \\ 
\textbf{Spectral Norm Bound} & 6.0 (WRN), 0.95 (BERT)  & 
Length-scale Parameter &  2.0  \\ 
 &  & $L_2$ Regularization & 0.0   \\ 
 &  & \textbf{Covariance Ridge Factor} & 0.001   \\ 
 &  & \textbf{Covariance Discount Factor} & 0.999   \\ 
 & & Projected Input Dimension & 128 (WRN) None (BERT) \\ 
 & & Input Layer Normalization & True \\
\bottomrule
\end{tabular}
}
\caption{Hyperparameters of SNGP used in the experiments, where important hyperparameters are highlighted in bold.}
\end{table}

\paragraph{Data Preparation and Computing Infrastructure} For CIFAR-10 and CIFAR-100, we followed the original Wide ResNet work to apply the standard data augmentation (horizontal flips and random crop-ping with 4x4 padding) and used the same hyperparameter and training setup \citep{zagoruyko_wide_2017}. The only exception is the learning rate and training epochs, where we find a smaller learning rate ($0.04$ for CIFAR-10 and $0.08$ for CIFAR100, v.s. $0.1$ for the original WRN model) and longer epochs ($250$ for SNGP v.s. $200$ for the original WRN model) leads to better performance.

For CLINC OOS intent understanding data, we pre-tokenized the sentences using the standard BERT tokenizer\footnote{\url{https://github.com/google-research/bert}} with maximum sequence length 32, and created standard binary input mask for the BERT model that returns 1 for valid tokens and 0 otherwise. Following the original BERT work, we used the Adam optimizer with  weight decay rate $0.01$ and warmup proportion $0.1$. We initialize the model from the official BERT$_{\texttt{Base}}$ checkpoint\footnote{\url{https://storage.googleapis.com/bert_models/2020_02_20/uncased_L-12_H-768_A-12.zip}}.
For this fine-tuning task, we using a smaller step size ($5e-5$ for SNGP .v.s. $1e-4$ for the original BERT model) but shorter epochs ($40$ for SNGP v.s. $150$ for the original BERT model) leads to better performance. When using spectral normalization, we set the hyperparameter $c=0.95$ and apply it to the pooler dense layer of the classification token. We do not spectral normalization to the hidden transformer layers, as we find the pre-trained BERT representation is already competent in preserving input distance due to the masked language modeling training, and further regularization may in fact harm its predictive and calibration performance. 

All models are implemented in TensorFlow and are trained on 8-core Cloud TPU v2 with 8 GiB of high-bandwidth memory (HBM) for each TPU core. We use batch size 32 per core.

\paragraph{Evaluation}
For CIFAR-10 and CIFAR-100, we evaluate the model's predictive accuracy and calibration error under both clean and corrupted versions of the CIFAR testing data. The corrupted data, termed CIFAR10-C, includes 15 types of corruptions, e.g., noise, blurring, pixelation, etc, over 5 levels of corruption intensity  \citep{hendrycks_benchmarking_2018}. We also evaluate the model performance in \gls{OOD} detection by using the CIFAR-10/CIFAR-100 model's uncertainty estimate as a predictive score for \gls{OOD} classification, where we consider a standard \gls{OOD} task by testing CIFAR-10/CIFAR-100 model's ability in detecting samples from the \glsfirst{SVHN} dataset \citep{netzer_reading_2011}, and a more difficult \gls{OOD} task by testing CIFAR-10's ability in detecting samples from the CIFAR-100 dataset, and vice versa. Specifically, for all models, we compute the OOD uncertainty score using the so-called \textit{Dempster-Shafer metric} \citep{sensoy_evidential_2018}, which empirically leads to better performance for a distance-aware model. Given logits for $K$ classes  $\{h_k(\bx_i)\}_{k=1}^K$, this metric computes its uncertainty for a test example $\bx_i$ as:
\begin{align}
    u(\bx_i) = \frac{K}{K + \sum_{k=1}^K \exp\big(h_k(\bx_i)\big)}.
\end{align}
As shown, for a distance-aware model where the magnitude of the logits reflects the distance from the observed data manifold, $u(\bx_i)$ can be a more effective metric since it is monotonic to the magnitude of the logits. On the other hand, the maximum probability $p_{\texttt{max}}=\arg\max_k \; \exp(h_k)/\sum_{k=1}^K \exp\big(h_k(\bx_i)\big)$ does not take advantage of this information since it normalizes over the exponentiated logits.

In terms of evaluation metrics, we assess the model's calibration performance using the empirical estimate of \gls{ECE}: $\hat{ECE}=\sum_{m=1}^M \frac{|B_m|}{n} |acc(B_m) - conf(B_m)|$ which estimates the difference in model's accuracy and confidence by partitioning model prediction into $M$ bins $\{B_m\}_{m=1}^M$ \citep{guo_calibration_2017}. In this work, we choose $M=15$. We assess the model's \gls{OOD} performance using \gls{AUPR}. Finally, we measure each method's inference latency by millisecond per image.

For CLINC OOS intent detection data, we evaluate the predictive accuracy on the in-domain test data, evaluate the ECE and \gls{OOD} detection performance on the combined in-domain and out-of-domain testing data, and we measure inference latency by millisecond per sentence.

\newpage
\section{Additional Related Work} 
\label{sec:related_app}

\textbf{Distance-preserving neural networks and bi-Lipschitz condition}
The theoretic connection between distance preservation and the bi-Lipschitz condition is well-established \citep{searcod_metric_2006}, and learning an approximately isometric, distance-preserving transform has been an important goal in the fields of dimensionality reduction \citep{blum_random_2006, perrault-joncas_metric_2017},  generative modeling \citep{lawrence_local_2006, dinh_nice:_2014, dinh_density_2016, jacobsen_i-revnet_2018}, and adversarial robustness \cite{jacobsen_excessive_2018, ruan_reachability_2018, sokolic_robust_2017, tsuzuku_lipschitz-margin_2018, weng_evaluating_2018}. This work is a novel application of the distance preservation property for uncertainty quantification. There existing several methods for controlling the Lipschitz constant of a \gls{DNN} (e.g., gradient penalty or norm-preserving activation \citep{an_how_2015, anil_sorting_2019, chernodub_norm-preserving_2017, gulrajani_improved_2017}), and we chose spectral normalization in this work due to its simplicity and its minimal impact on a \gls{DNN}'s architecture and the optimization dynamics \citep{bartlett_representing_2018, behrmann_invertible_2019, rousseau_residual_2020}. 

\textbf{Open Set Classification} The uncertainty risk minimization problem in Section \ref{sec:theory} assumes a data-generation mechanism similar to the \textit{open set recognition} problem \citep{scheirer_probability_2014}, where the whole input space is partitioned into known and unknown domains. However, our analysis is unique in that it focuses on measuring a model's behavior in uncertainty quantification and takes a rigorous, decision-theoretic approach to the problem. As a result, our analysis works with a special family of risk functions (i.e., \textit{the strictly proper scoring rule}) that measure a model's performance in uncertainty calibration. Furthermore, it handles the existence of unknown domain via a minimax formulation, and derives the solution by using a generalized version of maximum entropy theorem for the Bregman scores \citep{grunwald_game_2004, landes_probabilism_2015}. The form of the optimal solution we derived in (\ref{eq:minimax_solution}) takes an intuitive form, and has been used by many empirical work as a training objective to leverage adversarial training and generative modeling to detect \gls{OOD} examples \citep{hafner_reliable_2018, harang_principled_2018, lee_training_2018, malinin_prior_2018, meinke_towards_2020}. Our analysis provide strong theoretical support for these practices in verifying rigorously the uniqueness and optimality of this solution, and also provides a conceptual unification of the notion of calibration and the notion of \gls{OOD} generalization. Furthermore, it is used in this work to motivate a design principle (\textit{input distance awareness}) that enables strong \gls{OOD} performance in discriminative classifiers without the need of explicit generative modeling.

\newpage
\section{Proof}
\subsection{Proof of Proposition \ref{thm:resnet_lipschitz}}
\label{sec:resnet_lipschitz_proof}

The proof for Proposition \ref{thm:resnet_lipschitz} is an adaptation of the classic result of \citep{bartlett_representing_2018} to our current context:

\begin{proof}
First establish some notations. We denote $I(\bx)=\bx$ the identity function such that for $h(\bx)=\bx + g(\bx)$, we can write $g = h - I$. 
For $h: \Xsc \rightarrow \Hsc$, denote $||h||= \sup \Big\{ \frac{||f(\bx)||_H}{||\bx||_X} \mbox{ for } \bx \in \Xsc, ||\bx|| > 0 \Big\}$. Also denote the Lipschitz seminorm for a function $h$ as:
\begin{align}
    ||h||_L = \sup\bigg\{
    \frac{||h(\bx)-h(\bx')||_H}{||\bx - \bx'||_X} \quad \mbox{for} \quad 
    \bx, \bx' \in \Xsc, \bx \neq \bx'
    \bigg\}
\end{align}
It is worth noting that by the above definitions, for two functions $(\bx' - \bx): \Xsc \times \Xsc \rightarrow \Xsc$ and $(h(\bx) - h(\bx')): \Xsc \times \Xsc \rightarrow \Hsc$ who shares the same input space, the Lipschitz inequality can be expressed using the $||.||$ norm, i.e., $||h(\bx) - h(\bx')||_H \leq \alpha ||\bx - \bx'||_X$ implies $||h(\bx') - h(\bx)|| \leq \alpha ||\bx - \bx'||$, and vice versa.\\

Now assume $\forall l$, $||g_l||_L=||h_l-I||_L\leq \alpha < 1$. We will show  Proposition \ref{thm:resnet_lipschitz} by first showing:
\begin{align}
(1-\alpha)||\bx-\bx'|| \leq ||h_l(\bx)-h_l(\bx')|| \leq (1+\alpha)||\bx-\bx'||,
\label{eq:bilip_single_original}
\end{align}
which is the bi-Lipschitz condition for a single residual block.

First show the left hand side:
\begin{align*}
    ||\bx - \bx'||
    &\leq ||\bx  - \bx' - (h_l(\bx) - h_l(\bx')) + (h_l(\bx) - h_l(\bx'))||
    \nonumber \\
    &\leq ||(h_l(\bx')-\bx') - (h_l(\bx) - \bx)|| + ||h_l(\bx) - h_l(\bx')||
    \nonumber \\
    &\leq ||g_l(\bx') - g_l(\bx)|| + ||h_l(\bx) - h_l(\bx')||
    \nonumber \\
    &\leq \alpha||\bx' - \bx|| + ||h_l(\bx) - h_l(\bx')||,
\end{align*}
where the last line follows by the assumption $||g_l||_L \leq \alpha$. Rearranging, we get:
\begin{align}
    (1 - \alpha)||\bx - \bx'|| \leq ||h_l(\bx) - h_l(\bx')||.
    \label{eq:bilip_lhs}
\end{align}

Now show the right hand side:
\begin{align*}
    ||h_l(\bx) - h_l(\bx')|| 
    &= ||\bx + g_l(\bx) - (\bx' + g_l(\bx'))||
    \leq ||\bx  - \bx'|| + ||g_l(\bx) - g_l(\bx'))||
    \leq (1 + \alpha) ||\bx  - \bx'||.
    \label{eq:bilip_rhs}
\end{align*}
Combining (\ref{eq:bilip_lhs})-(\ref{eq:bilip_rhs}), we have shown (\ref{eq:bilip_single_original}), which also implies:
\begin{align}
(1-\alpha)||\bx-\bx'||_X \leq ||h_l(\bx)-h_l(\bx')||_H \leq (1+\alpha)||\bx-\bx'||_X
\end{align}

Now show the bi-Lipschitz condition for a $L$-layer residual network $h=h_L\circ h_{L-1} \circ \dots \circ h_1$. It is easy to see that by induction:
\begin{align}
(1-\alpha)^L||\bx-\bx'||_X \leq ||h(\bx)-h(\bx')||_H \leq (1+\alpha)^L||\bx-\bx'||_X
\end{align}
Denoting $L_1 = (1-\alpha)^L$ and $L_2=(1+\alpha)^L$, we have arrived at  expression in Proposition \ref{thm:resnet_lipschitz}.
\end{proof}

\subsection{Proof of Lemma \ref{thm:minimax_lemma}}
\label{sec:minimax_lemma_proof}
\begin{proof}
\input{section/proof/minimax_lemma}
\end{proof}

\end{document}

%% file: preamble/definition.tex
\newtheorem{theorem}{Theorem}
\newtheorem{lemma}{Lemma}

\newtheorem{proposition}{Proposition}

\newtheorem{definition}{Definition}

\makeatletter
\newcommand*{\addFileDependency}[1]{
  \typeout{(#1)}
  \@addtofilelist{#1}
  \IfFileExists{#1}{}{\typeout{No file #1.}}
}
\makeatother

%% file: preamble/acronym.tex
\newacronym{ID}{ID}{Integrated Dirichlet}
\newacronym{NGP}{NGP}{Normalized Gaussian Process}
\newacronym{CE}{CE}{Cross Entropy}

\newacronym{NLU}{NLU}{natural language understanding}
\newacronym{MCMC}{MCMC}{Markov chain Monte Carlo}
\newacronym{DST}{DST}{Dempster–Shafer theory}
\newacronym{bba}{bba}{basic belief assignment}
\newacronym{BMA}{BMA}{Bayesian model averaging}

\newacronym{DNN}{DNN}{deep neural network}

\newacronym{BNN}{BNNs}{Bayesian neural networks}
\newacronym{MCD}{MC Dropout}{Monte Carlo Dropout}
\newacronym{MCBN}{MCBN}{Monte Carlo Batch Normalization}
\newacronym{SWAG}{SWAG}{Stochastic Weight Averaging-Gaussian}

\newacronym{UQ}{UQ}{uncertainty quantification}

\newacronym{MAP}{MAP}{maximum a posterior}

\newacronym{CDF}{CDF}{cumulative distribution function}
\newacronym{PDF}{PDF}{probability density function}
\newacronym{PMF}{PMF}{probability mass function}
\newacronym{MGF}{MGF}{moment generating function}

\newacronym{GLM}{GLM}{generalized linear model}
\newacronym{GAM}{GAM}{generalized additive model}
\newacronym{MLE}{MLE}{maximum-likelihood}

\newacronym{HMC}{HMC}{Hamiltonian Monte Carlo}
\newacronym{VI}{VI}{Variational Inference}

\newacronym{CRPS}{CRPS}{\textit{continous ranked probability score}}
\newacronym{RMSE}{RMSE}{root mean squared error}
\newacronym{MSE}{MSE}{mean squared error}

\newacronym{KL}{KL}{Kullback-Leibler}
\newacronym{CvM}{CvM}{Cram\'{e}r von Mises}
\newacronym{ELBO}{ELBO}{\emph{evidence lower bound}}

\newacronym{BNE}{BNE}{\textit{Bayesian Nonparametric Ensemble}}
\newacronym{BAE}{BAE}{\textbf{Bayesian Additive Ensemble}}
\newacronym{stack}{stack}{\textbf{Bayesian Stacking}}
\newacronym{BME}{BME}{\textbf{Bayesian Mixture of Experts}}

\newacronym{CVI}{Calibrated VI}{\textbf{Calibrated Variational Inference}}
\newacronym{EnKF}{EnKF}{Ensemble Kalman Filters}

\newacronym{RKHS}{RKHS}{reproducing kernel Hilbert space}
\newacronym{RBF}{RBF}{radial basis function}
\newacronym{CI}{CI}{Coverage Index}

\newacronym[plural=Gaussian processes,firstplural=Gaussian processes (GPs)]{GP}{GP}{Gaussian process}
\newacronym{CGP}{CGP}{\textit{constrained Gaussian process}}

\newacronym{SGD}{SGD}{stochastic gradient descent}

\newacronym{ECE}{ECE}{Expected Calibration Error}
\newacronym{EB}{EB}{empirical Bayes}

\newacronym{DDGP}{DDGP}{deep Dirichlet-Gaussian process}
\newacronym{ARD}{ARD}{automatic relevance determination}

\newacronym{IND}{IND}{in-domain}
\newacronym{OOD}{OOD}{out-of-domain}

\newacronym{SNGP}{SNGP}{Spectral-normalized Neural Gaussian Process}
\newacronym{SVDKL}{SVDKL}{Stochastic Variational Deep Kernel Learning}

\newacronym{BERT}{BERT}{Bidirectional Encoder Representations from Transformers}

\newacronym{RFF}{RFF}{random Fourier feature}

\newacronym{SN}{SN}{spectral normalization}
\newacronym{DUQ}{DUQ}{Deterministic Uncertainty Quantification}
\newacronym{MCDGP}{MCD-GP}{Calibrated Deep Gaussian Process}

\newacronym{AUROC}{AUROC}{Area Under Receiver Operating Characteristic}
\newacronym{AUPR}{AUPR}{Area Under Precision-Recall}

\newacronym{BvM}{BvM}{Bernstein-von Mises}

\newacronym{SVHN}{SVHN}{Street View House Numbers}

\newacronym{ML}{ML}{machine-learning}

\newacronym{EP}{EP}{expectation propagation}

\newacronym{OOS}{OOS}{out-of-scope}

%% file: section/abstract.tex
\begin{abstract}
Bayesian neural networks 
and deep ensembles are principled approaches to estimate the predictive uncertainty of a deep learning model.  However their practicality in real-time, industrial-scale applications are limited due to their heavy memory and inference cost. This motivates us to study principled approaches to high-quality uncertainty estimation that require only a single  deep neural network (DNN).
By formalizing the uncertainty quantification as a minimax learning problem, we first identify \textit{distance awareness}, i.e., the model’s ability to properly quantify the distance of a testing example from the training data manifold, as a necessary condition for a DNN to achieve high-quality (i.e., minimax optimal) uncertainty estimation. We then propose \textit{\gls{SNGP}}, 
a simple method that improves the distance-awareness ability of modern DNNs, by adding a weight normalization step during training and replacing the output layer with a Gaussian Process. 
On a suite of vision and language understanding tasks and on modern architectures (Wide-ResNet and BERT), 
\gls{SNGP} is competitive with deep ensembles in prediction, calibration and out-of-domain detection, and outperforms the other single-model approaches.\footnote{Code available at  
\small{
\url{https://github.com/google/uncertainty-baselines/tree/master/baselines}.
}
} 
\end{abstract}

%% file: section/intro.tex
Efficient methods that reliably quantify a \gls{DNN}'s predictive uncertainty are important for industrial-scale, real-world applications, which include examples such as object recognition in autonomous driving \citep{feng_towards_2018}, ad click prediction in online advertising \citep{van_aken_challenges_2018}, and intent understanding in a conversational system \citep{zheng_out--domain_2020}.
For example, for a \gls{NLU} model
built for a domain-specific chatbot service (e.g, weather inquiry), the user's input utterance to the model can be of any topic, and the model needs to understand reliably and in real-time whether to abstain or to trigger one of its known APIs. 

When deep classifiers 
make predictions on input examples that are far from the support of the training set, their performance can be arbitrarily bad \citep{bartlett_classification_2008, cortes_learning_2009}. This motivates the need for methods that are aware of the distance between an input test example and previously seen training examples, so they can return a uniform (i.e., maximum entropy) distribution over output labels if the input is too far from the training set (i.e., the input is out-of-domain) \citep{hafner_reliable_2018}. \glspl{GP} with suitable kernels enjoy such a property. However, to apply \glspl{GP} to a high-dimensional machine learning problem, it is usually necessary to perform some form of feature extraction or dimensionality reduction 
using a \gls{DNN}. Ideally, the hidden representation of a \gls{DNN} should reflect a meaningful distance in the data manifold (e.g., the semantic textual similarity between two sentences), such that this ``distance aware" property is preserved.
However, as we will show in the experiments, this is often not guaranteed for common deep learning models (cf.~Figure \ref{fig:2d_exp}). 
\input{"./section/figures/2d_exp_moon"} 

We propose a simple solution to this problem, namely adding \textit{spectral normalization} to the weights in each (residual) layer \citep{miyato_spectral_2018}. We refer to our method as "Spectral-normalized Neural Gaussian Processes" (SNGP). 
We show that this provides bounds on $||h(\bx) - h(\bx')||_H$ relative to $||\bx-\bx'||_X$, where $\bx$ and $\bx'$ are two inputs, $h(\bx)$ is a deep feature extractor, and $||.||_X$ a semantically meaningful distance for the data manifold.  
We can then safely pass $h(\bx)$ into a distance-aware GP output layer. 
To ensure computational scalability, we approximate the GP posterior using a Laplace approximation to the random feature expansion of the GP, which gives rise to a model posterior that can be learned scalably and in closed-form with minimal modification to the training pipeline of a deterministic \gls{DNN}, and allows us to efficiently compute the predictive uncertainty on a per-input basis without Monte Carlo sampling.

In the rest of this paper, we first theoretically motivate the importance of \textit{distance awareness} for a model's ability uncertainty estimation by studying it as a minimax learning problem (Section \ref{sec:theory}). We then introduce our SNGP method in detail in Section \ref{sec:method}, and experimentally evaluate its performance against  other single-model approaches as well as deep ensembles in Section \ref{sec:exp} \citep{lakshminarayanan_simple_2017}. On two challenging real world problems, namely image classification (using a Wide Resnet model on CIFAR-10 and CIFAR-100) and conversational intent understanding (using a BERT model on CLINC \gls{OOS} intent dataset), 
we show that the \gls{SNGP} method attains an uncertainty performance (e.g., calibration and \gls{OOD} detection) that is competitive with that of a deep ensemble, while maintaining the accuracy and latency of a single deterministic \gls{DNN}.

%% file: section/figures/2d_exp_moon.tex
\begin{figure}[th!]
    \centering
    \subcaptionbox{Gaussian Process\label{fig:2d_exp_oval_gp}}{
    \includegraphics[width=0.18\columnwidth]{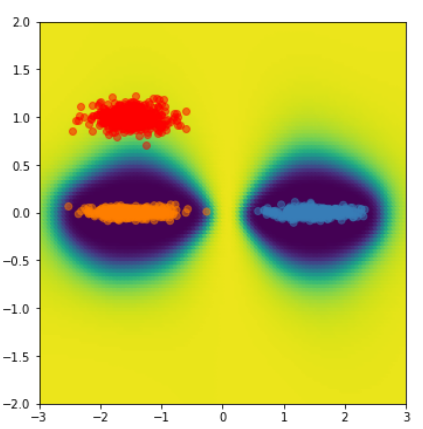}}
    \subcaptionbox{Deep Ensemble\label{fig:2d_exp_oval_deepens}}{
    \includegraphics[width=0.18\columnwidth]{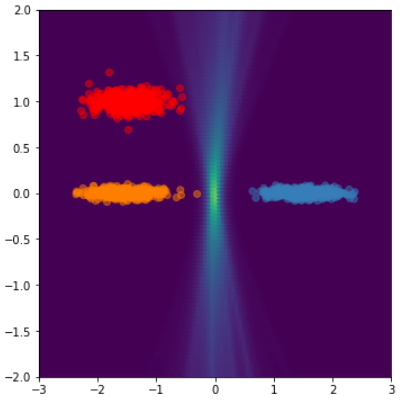}}
    \subcaptionbox{MC Dropout\label{fig:2d_exp_oval_mc}}{
    \includegraphics[width=0.18\columnwidth]{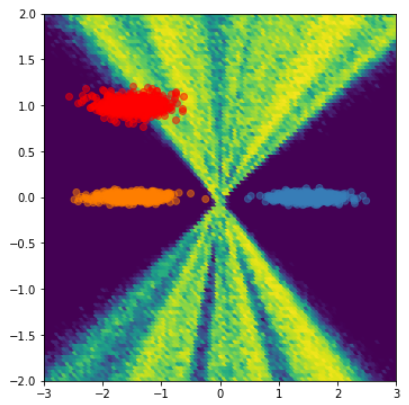}}
    \subcaptionbox{DNN-GP \label{fig:2d_exp_oval_deepgp}}{
    \includegraphics[width=0.18\columnwidth]{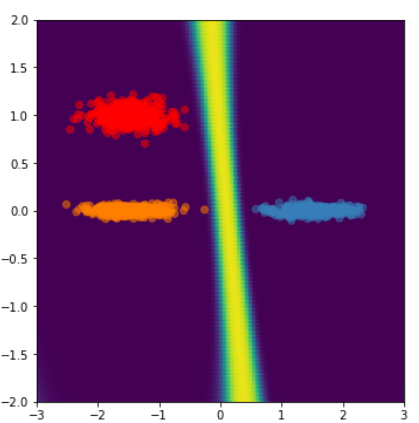}}
    \subcaptionbox{\gls{SNGP} (Ours)\label{fig:2d_exp_oval_sngp}}{
    \includegraphics[width=0.21\columnwidth]{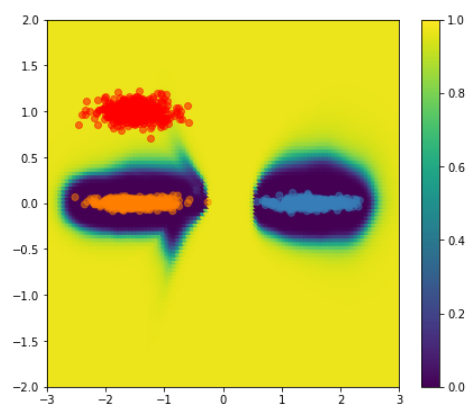}}

    \subcaptionbox{Gaussian Process\label{fig:2d_exp_moon_gp}}{
    \includegraphics[width=0.18\columnwidth]{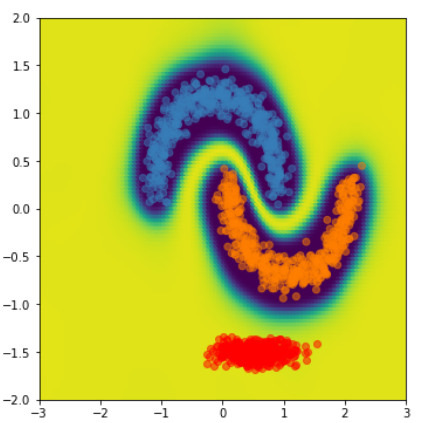}}
    \subcaptionbox{Deep Ensemble\label{fig:2d_exp_moon_deepens}}{
    \includegraphics[width=0.18\columnwidth]{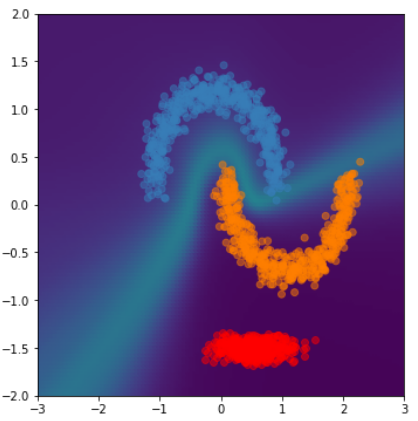}}
    \subcaptionbox{MC Dropout\label{fig:2d_exp_moon_mc}}{
    \includegraphics[width=0.18\columnwidth]{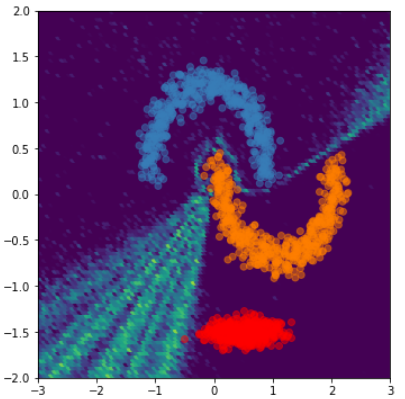}}
    \subcaptionbox{DNN-GP \label{fig:2d_exp_moon_deepgp}}{
    \includegraphics[width=0.18\columnwidth]{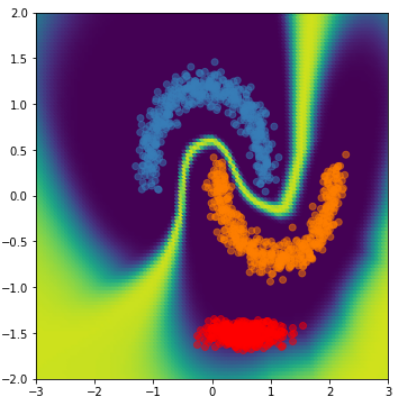}}
    \subcaptionbox{\gls{SNGP} (Ours)\label{fig:2d_exp_moon_sngp}}{
    \includegraphics[width=0.21\columnwidth]{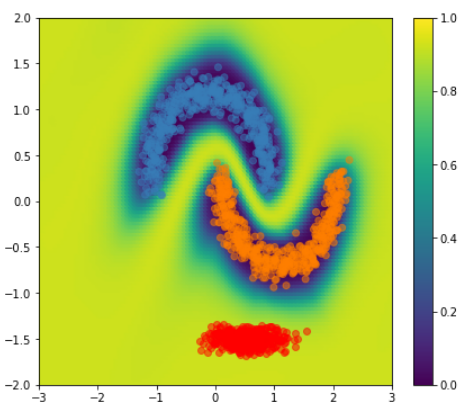}}
   \vspace{-0.5em}
    \caption{
    \small{
    The uncertainty surface of a \gls{GP} and different DNN approaches on the \textit{two ovals} (Top Row) and \textit{two moons} (Bottom Row) 2D classification benchmarks. SNGP is the only \gls{DNN}-based approach achieving a distance-aware uncertainty similar to the gold-standard \gls{GP}.
    Training data for positive ({\color{BurntOrange}{\textbf{Orange}}}) and negative classes ({\color{blue}\textbf{Blue}}). OOD data ({\color{red}\textbf{Red}}) not observed during training. Background color represents the  estimated model uncertainty (See \ref{fig:2d_exp_oval_sngp} and \ref{fig:2d_exp_moon_sngp} for color map).
    See Section 5.1 for details.
    }}
     \vspace{-0.9em}
    \label{fig:2d_exp}
\end{figure}

%% file: section/theory.tex
\paragraph{Notation and Problem Setup}
Consider a data-generation distribution $p^*(y|\bx)$, where $y \in \{1, \dots, K\}$ is the space of $K$-class labels, and $\bx \in \Xsc \subset \real^d$ is the input data manifold 
equipped with a suitable metric $||.||_X$. 
In practice, the training data $\Dsc=\{y_i, \bx_i\}_{i=1}^N$ is often collected from a subset of the full input space $\Xsc_{\texttt{IND}} \subset \Xsc$. As a result, the full data-generating distribution $p^*(y|\bx)$ is in fact a mixture of an  \gls{IND} distribution $p_{\texttt{IND}}(y|\bx) = p^*(y|\bx, \bx \in \Xsc_{\texttt{IND}})$ and also an \gls{OOD} distribution $p_{\texttt{OOD}}(y|\bx)=p^*(y|\bx, \bx \not\in \Xsc_{\texttt{IND}})$ \citep{meinke_towards_2020, scheirer_probability_2014}:
\begin{alignat}{4}
    p^*(y |\bx) 
    &= \qquad\qquad 
    p^*(y, \bx \in \Xsc_{\texttt{IND}}|\bx)
    && + \qquad\qquad
    p^*(y, \bx \not\in \Xsc_{\texttt{IND}}|\bx)
    \nonumber\\
    &= p^*(y|\bx, \bx \in \Xsc_{\texttt{IND}}) * p^*(\bx \in \Xsc_{\texttt{IND}}) 
    && + p^*(y|\bx, \bx \not\in \Xsc_{\texttt{IND}}) * p^*(\bx \not\in \Xsc_{\texttt{IND}}).
    \label{eq:true_dist}
\end{alignat}
During training, the model learns the in-domain distribution $p^*(y|\bx, \bx \in \Xsc_{\texttt{IND}})$ from the data $\Dsc$, but does not have knowledge about $p^*(y|\bx, \bx \not\in \Xsc_{\texttt{IND}})$.
In the weather-service chatbot example, the out-of-domain space $\Xsc_{\texttt{OOD}} = \Xsc / \Xsc_{\texttt{IND}}$ is the space of all natural utterances not related to weather queries, whose elements usually do not have a meaningful correspondence with the in-domain intent labels $y_k \in \{1, \dots, K\}$. 
Therefore, the out-of-domain distribution $p^*(y|\bx, \bx \not\in \Xsc_{\texttt{IND}})$ can be very different from the in-domain distribution $p^*(y|\bx, \bx \in \Xsc_{\texttt{IND}})$, and we only expect the model to generalize well within $\Xsc_{\texttt{IND}}$. However, during testing, the model needs to construct a predictive distribution $p(y|\bx)$ for the entire input space $\Xsc=\Xsc_{\texttt{IND}} \cup \Xsc_{\texttt{OOD}}$, since the users' utterances can be of any topic.




\vspace{-0.5em}
\subsection{Uncertainty Estimation as a Minimax Learning Problem} 
\label{sec:minimax}

To formulate the uncertainty estimation as a learning problem under (\ref{eq:true_dist}), we need to define a loss function to measure a model $p(y|\bx)$'s quality of predictive uncertainty. A popular uncertainty metric is \gls{ECE}, defined as $C(p, p^*) = E\big[ |E(y^*=\hat{y}|\hat{p}=p) - p | \big]$, which measures the difference in expectation between the model's predictive confidence (e.g., the maximum probability score) and its actual accuracy \citep{guo_calibration_2017,nixon2019measuring}. However, \gls{ECE} is not suitable as a loss function, since it is not uniquely minimized at $p = p^*$.
Specifically, there can exist a trivial 
predictor that ignores the input example and achieves perfect calibration by predicting randomly according to the marginal distribution of the labels \citep{gneiting_probabilistic_2007}.

To this end, a theoretically more well-founded 
uncertainty metric is to examine \textit{strictly proper scoring rules} \citep{gneiting_strictly_2007} $s(.,p^*)$, which are uniquely minimized by the true distribution $p=p^*$.
Examples include log-loss and Brier score. Proper scoring rules are related to \gls{ECE} in that each is an upper bound of the calibration error by the classic calibration-refinement decomposition   \citep{brocker2009reliability}.  Therefore, minimizing a proper scoring rule implies minimizing the calibration error of the model.
Consequently, we can formalize the problem of uncertainty quantification as the problem of constructing an optimal predictive distribution $p(y|\bx)$ to minimize the expected risk over the entire $\bx \in \Xsc$, i.e., an \textit{Uncertainty Risk Minimization} problem:
\begin{align}
    \inf_{p \in \Psc} S(p, p^*) = 
    \inf_{p \in \Psc} \underset{\bx \in \Xsc}{E}
    \big[s(p, p^*|\bx) \big].
    \label{eq:brier_risk}
\end{align}
Unfortunately, directly minimizing (\ref{eq:brier_risk}) over the entire input space $\Xsc$ is not possible even with infinite amounts of data. This is because since the data is  collected only from $\Xsc_{\texttt{IND}}$, the true \gls{OOD} distribution $p^*(y|\bx, \bx \not\in \Xsc_{\texttt{IND}})$ is never learned by the model, and generalization is not guaranteed since $p^*(y|\bx, \bx \in \Xsc_{\texttt{IND}})$ and $p^*(y|\bx, \bx \not\in \Xsc_{\texttt{IND}})$ are not assumed to be similar. As a result, the naive practice of using a model trained only with in-domain data to generate \gls{OOD} predictions can lead to arbitrarily bad results, since nature can happen to produce an \gls{OOD} distribution $p^*(y|\bx, \bx \not\in \Xsc_{\texttt{IND}})$ that is at odds with the model prediction. 
This is clearly undesirable for safety-critical applications. 
To this end, a more prudent strategy is to instead
minimize the \textit{worst-case} risk with respect to all possible $p^* \in \Psc^*$, i.e., construct $p(y|x)$ to minimize the \textit{Minimax Uncertainty Risk}:
\begin{align}
\underset{p \in \Psc}{\operatorname{inf}}  \Big[ \sup_{p^* \in \Psc^*} \; S(p, p^*) \Big].
\label{eq:minimax_loss}
\end{align}
In game-theoretic nomenclature, the uncertainty estimation problem acts as a two-player game of model v.s. nature, where the goal of the model is to produce a minimax strategy $p$ that minimizes the risk $S(p, p^*)$ against all possible (even adversarial) moves $p^*$ of nature.
Under the classification task and for Brier score, the solution to the minimax problem (\ref{eq:minimax_loss}) adopts a simple and elegant form: 
\begin{align}
    p(y |\bx) 
    &= 
    p(y|\bx, \bx \in \Xsc_{\textup{\texttt{IND}}}) * p^*(\bx \in \Xsc_{\textup{\texttt{IND}}}) 
    + 
    p_{\textup{\texttt{uniform}}}(y|\bx, \bx \not\in \Xsc_{\textup{\texttt{IND}}}) * p^*(\bx \not\in \Xsc_{\textup{\texttt{IND}}}).
    \label{eq:minimax_solution}
\end{align}
This is very intuitive: if an input point is in the training data domain, trust the model, otherwise use a uniform (maximum entropy) prediction.
For the practice of uncertainty estimation, (\ref{eq:minimax_solution}) is conceptually important in that it verifies that 
there exists a unique optimal solution to the uncertainty estimation problem (\ref{eq:minimax_loss}). Furthermore, this optimal solution can be constructed conveniently. Specifically, it can be constructed as a mixture of a discrete uniform distribution $p_{\texttt{uniform}}$ and the in-domain predictive distribution $p(y|\bx, \bx \in \Xsc_{\textup{\texttt{IND}}})$ that the model has already learned from data, \textit{assuming one can quantify $p^*(\bx \in \Xsc_{\textup{\texttt{IND}}})$ well}. 
In fact, the expression (\ref{eq:minimax_solution}) can be shown to be optimal for a broad family of scoring rules known as the Bregman score, which includes the Brier score and the widely used \textit{log score} as the special cases. We derive (\ref{eq:minimax_solution}) in 
Appendix \ref{sec:minimax_app}.


\vspace{-0.5em}
\subsection{Input Distance Awareness as a Necessary Condition}
\label{sec:da}

In light of Equation (\ref{eq:minimax_solution}), a key capacity for a deep learning model to reliably estimate predictive uncertainty 
is its ability to quantify, either explicitly or implicitly, the domain probability $p(\bx \in \Xsc_{\textup{\texttt{IND}}})$.
This requires the model to have a good notion of the distance (or dissimilarity) between a testing example $\bx$ and the training data $\Xsc_{\texttt{IND}}$ with respect to a \textit{meaningful} distance $||.||_X$ for the data manifold (e.g.,  semantic textual similarity \citep{cer_semeval-2017_2017} for language data). Definition \ref{def:da} makes this notion more precise:
\begin{definition}[Input Distance Awareness] Consider a predictive distribution $p(y|\bx)$ trained on a domain $\Xsc_{\textup{\texttt{IND}}} \subset \Xsc$, where $(\Xsc, ||.||_X)$ is the input data manifold equipped with a suitable metric $||.||_X$. We say $p(y|\bx)$ is \underline{input distance aware} if there exists $u(\bx)$ a summary statistic of $p(y|\bx)$ that quantifies model uncertainty  (e.g., entropy, predictive variance, etc) that reflects the distance between $\bx$ and the training data with respect to $||.||_X$, i.e.,
$$u(\bx) = v\big(
d(\bx, \Xsc_{\textup{\texttt{IND}}})\big)$$
where $v$ is a monotonic function and $d(\bx, \Xsc_{\textup{\texttt{IND}}}) = E_{\bx' \sim \Xsc_{\textup{\texttt{IND}}}} ||\bx - \bx'||^2_X$.
is the distance between $\bx$ and the training data domain. 
\label{def:da}
\end{definition}
A classic model that satisfies the \textit{distance-awareness} property is a \glsfirst{GP} with a \gls{RBF} kernel. Its predictive distribution $p(y|\bx)= softmax(g(\bx))$ is a softmax transformation of the \gls{GP} posterior $g \sim GP$ under the cross-entropy likelihood, and its predictive uncertainty can be expressed by the posterior variance $u(\bx^*)=var(g(\bx^*)) = 1 - \bk^{*\top}\bV\bk^*$ for $\bk^*_i=exp(-\frac{1}{2 l}||\bx^* - \bx_i||_X^2)$ and $\bV_{N \times N}$  a fixed matrix determined by data. Then $u(\bx^*)$ increases monotonically toward 1 as $\bx^*$ moves further away from $\Xsc_{\texttt{IND}}$ \citep{rasmussen_gaussian_2006}.
In view of the expression (\ref{eq:minimax_solution}), the \textit{input distance awareness} property is important for both calibration and \gls{OOD} detection. However, this property is not guaranteed for a typical deep learning model \citep{hein_why_2019}. Consider a discriminative deep classifier with dense output layer $logit_k(\bx) = h(\bx)^\top \bbeta_k$, whose model confidence (i.e., maximum predictive probability) is characterized by the magnitude of the class logits, which is defined by the inner product distances between the hidden representation $h(\bx)$ and the decision boundaries $\{\bbeta_k\}_{k=1}^K$ (see, e.g., Figure \ref{fig:2d_exp_oval_deepens}-\ref{fig:2d_exp_oval_mc} and \ref{fig:2d_exp_moon_deepens}-\ref{fig:2d_exp_moon_mc}). As a result, the model computes confidence for a $\bx^*$ based not on its distance from the training data $\Xsc_{\texttt{IND}}$, but based on its distance from the decision boundaries, i.e., the model uncertainty is not \textit{input distance aware}. 

\textbf{Two Conditions for Input Distance Awareness in Deep Learning} 
Notice that a deep learning model $logit(\bx) = g \circ h(\bx)$ is commonly composed of a hidden mapping $h: \Xsc \rightarrow \Hsc$  that maps the input $\bx$ into a hidden representation space $h(\bx) \in \Hsc$, and an output layer $g$ that maps $h(\bx)$ to the label space. To this end, a \gls{DNN}  $logit(\bx) = g \circ h(\bx)$ can be made \textit{input distance aware} 
via a combination of two conditions: 
\textbf{(1)} make the output layer $g$
\textit{distance aware}, so it outputs 
an uncertainty metric reflecting distance in the hidden space $||h(\bx) - h(\bx')||_H$ (in practice, this can be achieved by using a \gls{GP} with a shift-invariant kernel as the output layer),
and \textbf{(2)} make the hidden mapping \textit{distance preserving} (defined below), so that the distance in the hidden space $||h(\bx) - h(\bx')||_H$ has a meaningful correspondence to the distance $||\bx - \bx'||_X$ in the data manifold. From the mathematical point of view, this is equivalent to requiring $h$ to satisfy the \textit{bi-Lipschitz} condition \citep{searcod_metric_2006}:
\begin{align}
    L_1 * ||\bx_1 - \bx_2||_X \leq ||h(\bx_1) - h(\bx_2)||_H \leq L_2 * ||\bx_1 - \bx_2||_X,
    \label{eq:dp}
\end{align}
for positive and bounded constants $0 < L_1 < 1 < L_2$. It is worth noticing that for a deep learning model, the bi-Lipschitz condition (\ref{eq:dp}) usually leads the model's hidden space to preserve a semantically meaningful distance in the input data manifold $\Xsc$, rather than a naive metric such as the square distance in the pixel space. This is because that the upper Lipschitz bound $||h(\bx_1) - h(\bx_2)||_H \leq L_2 * ||\bx_1 - \bx_2||_X$ is an important condition for the adversarial robustness of a deep network, which prevents the hidden representations $h(\bx)$ from being overly sensitive to the semantically meaningless perturbations in the pixel space \citep{ruan_reachability_2018, weng_evaluating_2018, tsuzuku_lipschitz-margin_2018, jacobsen_excessive_2018, sokolic_robust_2017}. On the other hand, 
the lower Lipschitz bound $||h(\bx_1) - h(\bx_2)||_H \geq L_1 * ||\bx_1 - \bx_2||_X$ prevents the hidden representation from being unnecessarily invariant to the semantically meaningful changes in the input manifold \citep{jacobsen_exploiting_2019, van_amersfoort_simple_2020}. Combined together, the bi-Lipschitz condition essentially encourages $h$ to be an approximately isometric mapping, 
thereby ensuring that the learned representation $h(\bx)$ has a robust and meaningful correspondence with the semantic properties of the input data $\bx$.
Although not stated explicitly, learning an approximately isometric and geometry-preserving mapping is a common goal in machine learning. For example, image classifiers strive to learn a mapping from image manifold to a hidden space that can be well-separated by a set of linear decision boundaries, and sentences encoders aim to project sentences into a vector space where the cosine distance reflects the semantic similarity in natural language. Finally, it is worth noting that preserving such approximate isometry in a neural network is possible even after significant dimensionality reduction \citep{blum_random_2006, hauser_principles_2017, perrault-joncas_metric_2017, rousseau_residual_2020}.




%% file: section/method.tex

In this section we propose \textit{\glsfirst{SNGP}}, a simple method to improve the \textit{input distance awareness} ability of a modern residual-based \gls{DNN} (e.g., ResNet, Transformer) by (1) making the output layer \textit{distance aware} and (2) making the hidden layers \textit{distance preserving}, as discussed in Section \ref{sec:da}. Full method is summarized in Algorithms \ref{alg:training}-\ref{alg:prediction}.

\vspace{-0.5em}
\subsection{Distance-aware Output Layer via Laplace-approximated Neural Gaussian Process}
\label{sec:gp}

To make the output layer $g: \Hsc \rightarrow \Ysc$ distance aware, \gls{SNGP} replaces the typical dense output layer with a \glsfirst{GP} with an \gls{RBF} kernel, whose posterior variance at $\bx^*$ is characterized by its $L_2$ distance from the training data in the hidden space. 
Specifically, given $N$ training samples $\Dsc=\{y_i, \bx_i\}_{i=1}^N$ and denoting $h_i=h(\bx_i)$, the Gaussian-process output layer $\bg_{N \times 1} = [g(h_1), \dots, g(h_N)]^\top$ follows a multivariate normal distribution \textit{a priori}:    
\begin{align}
    \bg_{N \times 1} \sim MVN(\bzero_{N \times 1}, \bK_{N \times N}), 
  \mbox{where} \  
    \bK_{i, j} = \exp(-||h_i - h_j||_2^2/2),
    \label{eq:gp_prior}
\end{align}
and the posterior distribution is computed as $p(g|\Dsc) \propto p(\Dsc|g)p(g)$ where $p(g)$ is the \gls{GP} prior in (\ref{eq:gp_prior}) 
and $p(\Dsc|g)$ is the data likelihood for classification (i.e., the exponentiated cross-entropy loss). However, computing the exact Gaussian process posterior for a large-scale classification task is both analytically intractable and computationally expensive,
In this work, we propose a simple approximation strategy for \gls{GP} that is based on a Laplace approximation to the \gls{RFF} expansion of the \gls{GP} posterior \citep{rasmussen_gaussian_2006}. Our approach gives rise to a closed-form posterior that is end-to-end trainable with the rest of the neural network, and empirically leads to an improved quality in estimating the posterior uncertainty. Specifically, we first approximate the \gls{GP} prior in (\ref{eq:gp_prior}) by deploying a low-rank approximation to the kernel matrix $\bK=\bPhi\bPhi^\top$ using random features \citep{rahimi_random_2008}:
\begin{align}
    \bg_{N \times 1} \sim MVN(\bzero_{N \times 1}, \bPhi\bPhi^\top_{N \times N}), 
   \qquad \mbox{where} \qquad 
    \bPhi_{i, D_L \times 1} = 
    \sqrt{2/D_L} * 
    \cos(-\bW_L h_i + \bb_L),
    \label{eq:rff_gp}
\end{align}
where $h_i=h(\bx_i)$ is the hidden representation in the penultimate layer with dimension $D_{L-1}$. $\bPhi_i$ is the final layer with dimension $D_L$, it contains $\bW_{L, D_L \times D_{L-1}}$ a fixed weight matrix whose entries are sampled i.i.d. from $N(0, 1)$, and $\bb_{L, D_L \times 1}$ a fixed bias term whose entries are sampled i.i.d.~from $Uniform(0, 2\pi)$. As a result, for the $k^{th}$ logit, the \gls{RFF} approximation to the \gls{GP} prior in (\ref{eq:gp_prior}) can be written as a neural network layer with fixed hidden weights $\bW$ and learnable output weights $\bbeta_k$:
\begin{align}
    g_k(h_i) &= \sqrt{2/D_L} * \cos(-\bW_L h_i + \bb_L)^\top \bbeta_k, 
    \qquad \mbox{with prior} \qquad 
    &\bbeta_{k, D_L \times 1} \sim N(0, \bI_{D_L \times D_L}).
    \label{eq:rff_lr}
\end{align}
Notice that conditional on $h$, $\beta=\{\bbeta_k\}_{k=1}^K$ is the only learnable parameter in the model. As a result, the \gls{RFF} approximation in  (\ref{eq:rff_lr}) reduces an infinite-dimensional \gls{GP} to a standard Bayesian linear model, for which many posterior approximation methods (e.g., \gls{EP}) can be applied \citep{minka_family_2001}. In this work, we choose the Laplace method due to its simplicity and the fact that its posterior variance has a convenient closed form \citep{rasmussen_gaussian_2006}.
Briefly, the Laplace method approximates the \gls{RFF} posterior $p(\beta|\Dsc)$ using a Gaussian likelihood centered around the \gls{MAP} estimate $\hat{\beta} = \mbox{argmax}_{\beta}\,  p(\beta|\Dsc)$, such that $p(\beta_k | \Dsc) \approx MVN(\hat{\beta}_k, \hat{\bSigma}_{k}=\hat{\bH}^{-1}_{k})$, where $\hat{\bH}_{k, (i,j)} = \frac{\partial^2}{\partial \beta_i \partial \beta_j} \log \, p(\beta_k|\Dsc)|_{\beta_k=\hat{\beta}_k}$ is the $D_L \times D_L$ Hessian matrix of the log posterior likelihood evaluated at the \gls{MAP} estimates. Under the linear-model formulation of the \gls{RFF} posterior, the posterior precision matrix (i.e., the inverse covariance matrix) adopts a simple expression $\hat{\bSigma}^{-1}_k = 
\bI + \sum_{i=1}^N \hat{p}_{i,k}(1-\hat{p}_{i,k}) \Phi_i \Phi_i^\top$, where $p_{i,k}$ is the model prediction $softmax(\hat{g}_i)$ under the \gls{MAP} estimates $\hat{\beta}=\{\beta_k\}_{k=1}^K$  \citep{rasmussen_gaussian_2006}. To summarize, the Laplace posterior for \gls{GP} under the \gls{RFF} approximation is:
\begin{align}
    \beta_k | \Dsc \sim 
    MVN(\hat{\beta}_k, \hat{\bSigma}_k), 
    \quad \mbox{where} \quad 
    \hat{\bSigma}^{-1}_k = 
\bI + \sum_{i=1}^N \hat{p}_{i,k}(1-\hat{p}_{i,k}) \Phi_i \Phi_i^\top. 
\label{eq:gp_posterior}
\end{align}
During minibatch training, the posterior mean $\hat{\beta}$ is updated via regular \gls{SGD} with respect to the (unnormalized) log posterior $-\log p(\beta|\Dsc) = -\log p(\Dsc|\beta) + \frac{1}{2}||\bbeta||^2$ where $-\log p(\Dsc|\beta)$ is the cross-entropy loss. 
The posterior precision matrix is updated cheaply as 
$\hat{\bSigma}^{-1}_{k,t}=(1-m)*\hat{\bSigma}^{-1}_{k, t-1} + m*\sum_{i=1}^M \hat{p}_{i,k}(1-\hat{p}_{i,k})\Phi_i \Phi_i^\top$ for a minibatch of size $M$ and $m$ a small scaling coefficient. This computation only needs to be performed by passing through training data once at the final epoch.
As a result, the \gls{GP} posterior (\ref{eq:gp_posterior}) can be learned scalably and in closed-form with minimal modification to the training pipeline of a deterministic \gls{DNN}. 
It is worth noting that the Laplace approximation to the \gls{RFF} posterior is asymptotically exact by the virtue of the \gls{BvM} theorem and the fact that (\ref{eq:rff_lr}) is a finite-rank model \citep{dehaene_deterministic_2019, freedman_wald_1999, lecam_convergence_1973, panov_finite_2015}.


\vspace{-0.5em}
\subsection{Distance-preserving Hidden Mapping via Spectral Normalization}
\label{sec:sn}

Replacing the output layer $g$ with a Gaussian process only allows the model $logit(\bx) = g \circ h(\bx)$ to be aware of the distance in the hidden space $||h(\bx_1) - h(\bx_2)||_H$.
It is also important to ensure the hidden mapping $h$ is \textit{distance preserving} so that the distance in the hidden space $||h(\bx) - h(\bx')||_H$ has a meaningful correspondence to the distance in the input space $||\bx - \bx'||_X$. To this end, we notice that modern deep learning models (e.g., ResNets, Transformers) are commonly composed of residual blocks, i.e., $h(\bx)=h_{L-1} \circ \dots  \circ h_2  \circ h_1(\bx)$ where $h_l(\bx)= \bx + g_l(\bx)$. For such models, there exists a simple method to ensure  $h$ is \textit{distance preserving}: by bounding the Lipschitz constants of all nonlinear residual mappings $\{g_l\}_{l=1}^{L-1}$ to be less than 1. We state this result formally below: 
\begin{proposition}[Lipschitz-bounded residual block is distance preserving \citep{bartlett_representing_2018}]
Consider a hidden mapping $h: \Xsc \rightarrow \Hsc$ with residual architecture $h=h_{L-1} \circ \dots h_2 \circ h_1$ where $h_l(\bx) = \bx + g_l(\bx)$. If for  $0 < \alpha \leq 1$, all $g_l$'s are $\alpha$-Lipschitz, i.e., $||g_l(\bx) - g_l(\bx')||_H \leq \alpha ||\bx - \bx'||_X \quad \forall (\bx, \bx') \in \Xsc$. Then:
$$L_1 * ||\bx - \bx'||_X \leq || h(\bx) - h(\bx') ||_H \leq L_2 * ||\bx - \bx'||_X, $$
where $L_1 = (1-\alpha)^{L-1}$ and $L_2=(1+\alpha)^{L-1}$, i.e., $h$ is \textit{distance preserving}.
\label{thm:resnet_lipschitz}
\end{proposition}
Proof is in Appendix \ref{sec:resnet_lipschitz_proof}. The ability of a residual network to construct a geometry-preserving metric transform between the input space $\Xsc$ and the hidden space $\Hsc$ is well-established in learning theory and generative modeling literature, but the application of these results in the context of uncertainty estimation for \gls{DNN} appears to be new  \citep{bartlett_representing_2018, behrmann_invertible_2019, hauser_principles_2017,  rousseau_residual_2020}. 

Consequently, to ensure the hidden mapping $h$ is distance preserving, it is sufficient to ensure that the weight matrices for the nonlinear residual block $g_l(\bx)=\sigma(\bW_l\bx + \bb_l)$ to have spectral norm (i.e., the largest singular value) less than 1, since $||g_l||_{Lip} \leq ||\bW_l\bx + \bb_l||_{Lip} \leq ||\bW_l||_2 \leq 1$. In this work, we enforce the aforementioned Lipschitz constraint on $g_l$'s by applying the \textit{ \gls{SN}} on the weight matrices $\{\bW_l\}_{l=1}^{L-1}$ as recommended in \citep{behrmann_invertible_2019}. Briefly, at every training step, the \gls{SN} method first estimate the spectral norm $\hat{\lambda} \approx ||\bW_l||_2$ using the power iteration method \citep{gouk_regularisation_2018, miyato_spectral_2018}, and then normalizes the weights as:\\
\scalebox{0.9}{
\parbox{\linewidth}{%
\begin{align}
    \bW_l = 
    \begin{cases} 
    c * \bW_l / \hat{\lambda}  & 
    \mbox{if } c < \hat{\lambda}\\ 
    \bW_l & \mbox{otherwise}
    \end{cases}
    \label{eq:spec_norm}
\end{align}
}
}\\
where $c>0$ is a hyperparameter used to adjust the exact spectral norm upper bound on $||\bW_l||_2$ (so that $||\bW_l||_2 \leq c$). This hyperparameter is useful in practice since the other regularization mechanisms (e.g., Dropout, Batch Normalization) in the hidden layers can rescale the Lipschitz constant of the original residual mapping \citep{gouk_regularisation_2018}. Therefore, (\ref{eq:spec_norm}) allows us more flexibility in controlling the spectral norm of the neural network weights so it is the most compatible with the architecture at hand.\\
\textbf{Method Summary} We summarize the method in Algorithms  \ref{alg:training}-\ref{alg:prediction}. 
As shown, for every minibatch step, the model first updates the hidden-layer weights $\{\bW_l, \bb_l\}_{l=1}^{L-1}$ and the trainable output weights $\beta=\{\beta_k\}_{k=1}^K$ via \gls{SGD}, 
then performs spectral normalization, and finally (if in the final epoch) performs precision matrix update (Equation (\ref{eq:gp_posterior}). We discuss further details (e.g. computational complexity) in Appendix 
\ref{sec:method_sum}.
\input{section/figures/algorithm}

%% file: section/figures/algorithm.tex
\begin{center}
\scalebox{.85}{%
\begin{minipage}{0.5\textwidth}
\begin{algorithm}[H]
   \caption{\gls{SNGP} Training}
   \label{alg:training}
\begin{algorithmic}[1]
   \STATE {\bfseries Input:} \\
   Minibatches $\{D_i\}_{i=1}^N$ for $D_i=\{y_m, \bx_m\}_{m=1}^M$. 
   \vspace{0.2em}
   \STATE {\bfseries Initialize:} \\
    \vspace{-1.2em}
    $$\hat{\bSigma}=\bI, \bW_L \stackrel{iid}{\sim} N(0, 1), \bb_L \stackrel{iid}{\sim} U(0, 2\pi)$$.\\
     \vspace{-1.2em}
   \FOR{$\mathsf{train\_step}=1$ {\bfseries to} $\mathsf{max\_step}$}
   \STATE 
   \gls{SGD} update $\Big\{ \bbeta, \{\bW_l\}_{l=1}^{L-1}, \{\bb_l\}_{l=1}^{L-1} \Big\}$
   \STATE 
   Spectral Normalization $\{\bW_l\}_{l=1}^{L-1}$ 
   (\ref{eq:spec_norm}).
   \IF{$\mathsf{final\_epoch}$}
   \STATE 
   Update precision matrix $\{\hat{\bSigma}_{k}^{-1}\}_{k=1}^K$ 
   (\ref{eq:gp_posterior}).
   \ENDIF
   \ENDFOR
   \STATE Compute posterior covariance $\hat{\bSigma}_k=inv(\hat{\bSigma}_k^{-1})$.
\end{algorithmic}
\end{algorithm}
\end{minipage}

\hspace{1em}

\begin{minipage}{0.5\textwidth}
\begin{algorithm}[H]
   \caption{\gls{SNGP} Prediction}
   \label{alg:prediction}
\begin{algorithmic}[1]
   \STATE {\bfseries Input:} Testing example $\bx$.
   \STATE 
   Compute Feature: 
   \vspace{-0.5em}
   $$\bPhi_{D_L \times 1} = \sqrt{2/D_L} * \cos(\bW_L h(\bx) + \bb_L),$$ 
   \vspace{-1.2em}
   \STATE 
   Compute Posterior Mean: 
   \vspace{-0.5em}
   $$\logit_k(\bx)=\Phi^\top\bbeta_k$$ 
   \vspace{-1.2em}
   \STATE 
   Compute Posterior Variance: 
   \vspace{-0.5em}
   $$\var_k(\bx)=\Phi^\top\hat{\Sigma}_k\Phi.$$ 
   \vspace{-1em}
      \STATE 
      Compute Predictive Distribution:\\
   \vspace{-1.5em}
   $$p(y|\bx)=\int_{m \sim N(\logit(\bx), \var(\bx))}{\softmax(m)}$$ 
   \vspace{-1em}
\end{algorithmic}
\end{algorithm}
\end{minipage}
}
\end{center}

%% file: section/related.tex
\textbf{Single-model approaches to deep classifier uncertainty}
Recent work examines uncertainty methods that add few additional parameters or runtime cost to the base model. The state-of-the-art on large-scale tasks are efficient ensemble methods \citep{wen_batchensemble_2020,dusenberry_efficient_2020}, which cast a set of models under a single one, encouraging independent member predictions using low-rank perturbations.  These methods are parameter-efficient but still require multiple forward passes from the model. SNGP investigates an orthogonal approach that improves the  uncertainty quantification by imposing suitable regularization on a single model, and therefore requires only a single forward pass during inference.
There exists other runtime-efficient, single-model approaches to estimate predictive uncertainty, achieved by either replacing the loss function \citep{hein_why_2019, malinin_predictive_2018, malinin_prior_2018, sensoy_evidential_2018, shu_doc_2017}, the output layer \citep{bendale_towards_2016, tagasovska_single-model_2019, calandra_manifold_2016, macedo_isotropic_2020}, or computing a closed-form posterior for the output layer \citep{riquelme_deep_2018, snoek_scalable_2015, kristiadi_being_2020}. \gls{SNGP} builds on these approaches by also considering the intermediate representations which are necessary for good uncertainty estimation, and proposes a simple method (spectral normalization) to achieve it. A recent method named Deterministic Uncertainty Quantification (\textbf{DUQ}) also regulates the neural network mapping but uses a two-sided gradient penalty \citep{van_amersfoort_simple_2020}. The 
two-sided gradient penalty can be undesirable for a residual network, since imposing $||\nabla f||=1$ onto a residual connection $f(\bx)=\bx+g(\bx)$ can force $g(\bx)$ toward 0, leading to an identity mapping. We compare with DUQ in our experiments.

\textbf{Laplace approximation and GP inference with \gls{DNN}} Laplace approximation has a long history in \gls{GP} and NN literature \citep{tierney_approximate_1989, denker_transforming_1991, rasmussen_gaussian_2006, mackay_practical_1992, ritter_scalable_2018}, and the theoretical connection between a Laplace-approximated \gls{DNN} and \gls{GP} has being explored recently \citep{khan_approximate_2019}. Differing from these works, \gls{SNGP} applies the Laplace approximation to the posterior of a neural GP, rather than to a shallow GP or a dense-output-layer \gls{DNN}. Earlier works that combine a \gls{GP} with a \gls{DNN} usually perform MAP estimation  \citep{calandra_manifold_2016} or structured \gls{VI} \citep{bradshaw_adversarial_2017, wilson_stochastic_2016}. These approaches were shown to lead to poor calibration by recent work \citep{tran_calibrating_2019}, which proposed a simple fix by combing \gls{MCD} with random Fourier features, which we term Calibrated Deep Gaussian Process (\textbf{MCD-GP})\glsunset{MCDGP}.
\gls{SNGP} differs from \gls{MCDGP} in that it considered a different regularization approach (spectral normalization) and can compute its posterior uncertainty more efficiently in a single forward pass. We compare with \gls{MCDGP} in our experiments. 
Appendix \ref{sec:related_app} contains further related work on distance-preserving neural networks and open-set classification.

%% file: section/experiment.tex
\subsection{2D Synthetic Benchmark}
\label{sec:exp_2d}

We first study the behavior of the uncertainty surface of a \gls{SNGP} model under a suite of 2D classification benchmarks. 
Specifically, we consider the \textit{two ovals} benchmark (Figure \ref{fig:2d_exp}, row 1) and the \textit{two moons} benchmark (Figure \ref{fig:2d_exp}, row 2). The \textit{two ovals} benchmark consists of two near-flat Gaussian distributions, which represent the two in-domain classes (orange and blue) that are separable by a linear decision boundary. There also exists an \gls{OOD} distribution (red) that the model doesn't observe during training. Similarly, the \textit{two moons} dataset consists of two banana-shaped distributions separable by a nonlinear decision boundary. We consider a 12-layer, 128-unit deep architecture ResFFN-12-128. The full experimental details are in Appendix \ref{sec:exp_app}. 

Figure \ref{fig:2d_exp} shows the results, where the  background color visualizes the uncertainty surface output by each model. We first notice that the shallow Gaussian process models (Figures \ref{fig:2d_exp_oval_gp} and \ref{fig:2d_exp_moon_gp})  exhibit an expected behavior for high-quality predictive uncertainty: 
it generates low uncertainty in $\Xsc_{\texttt{IND}}$ that is supported by the training data (purple color), and generates high uncertainty when $\bx$ is far from $\Xsc_{\texttt{IND}}$ (yellow color), i.e., \textit{input distance awareness}. As a result, the shallow \gls{GP} model is able to assign low confidence to the \gls{OOD} data (colored in red), indicating reliable uncertainty quantification. On the other hand, deep ensembles (Figures \ref{fig:2d_exp_oval_deepens},  \ref{fig:2d_exp_moon_deepens}) and MC Dropout (Figures \ref{fig:2d_exp_oval_mc}, \ref{fig:2d_exp_moon_mc}) are based on dense output layers that are not distance aware. As a result, both methods quantify their predictive uncertainty based on the distance from the decision boundaries, assigning low uncertainty to \gls{OOD} examples even if they are far from the data.
Finally, the DNN-GP (Figures \ref{fig:2d_exp_oval_deepgp} and\ref{fig:2d_exp_moon_deepgp}) and \gls{SNGP} (Figures \ref{fig:2d_exp_oval_sngp} and\ref{fig:2d_exp_moon_sngp}) both use \gls{GP} as their output layers, but with \gls{SNGP} additionally imposing the spectral normalization on its hidden mapping $h(.)$. As a result, the DNN-GP's uncertainty surfaces are still strongly impacted by the distance from decision boundary, likely caused by the fact that the un-regularized hidden mapping $h(\bx)$ is free to discard information that is not relevant for prediction. On the other hand, the \gls{SNGP} is able to maintain the \textit{input distance awareness} property via its bi-Lipschitz constraint, and exhibits a uncertainty surface that is analogous to the gold-standard model (shallow \gls{GP}) despite the fact that \gls{SNGP} is based on a 12-layer network.

\vspace{-0.5em}

\subsection{Vision and Language Understanding}
\label{sec:exp_ml}
\paragraph{Baseline Methods} All methods included in the vision and language understanding experiments are summarized in Table \ref{tb:methods}. Specifically, we evaluate \gls{SNGP} on a Wide ResNet 28-10 \citep{zagoruyko_wide_2017} for image classification, and BERT$_{\texttt{base}}$ \citep{devlin_bert:_2018} for language understanding. We compare against a deterministic baseline and two ensemble approaches:
\textbf{\gls{MCD}} (with 10 dropout samples) and \textbf{deep ensembles} (with 10 models), all trained with a dense output layer and no spectral regularization. We consider three single-model approaches: \textbf{\gls{MCDGP}} (with 10 samples), \textbf{\gls{DUQ}} (see Section \ref{sec:related}). For all models that use \gls{GP} layer, we keep $D_L=1024$ and compute predictive distribution by performing Monte Carlo averaging with 10 samples. We also include two ablated version of \gls{SNGP}: \textbf{DNN-SN} which uses spectral normalization on its hidden weights and a dense output layer (i.e. distance preserving hidden mapping without distance-aware output layer), and \textbf{DNN-GP} which uses the \gls{GP} as output layer but without spectral normalization on its hidden layers (i.e., distance-aware output layer without distance-preserving hidden mapping). Further experiment details 
and recommendations for practical implementation
are in Appendix \ref{sec:exp_app}.  
All baselines are built on the \href{https://github.com/google/uncertainty-baselines}{\texttt{uncertainty\_baselines}} framework. 

\input{section/tables/methods}
\vspace{-1em}

\textbf{CIFAR-10 and CIFAR-100}
We evaluate the model's predictive accuracy and calibration error under both clean CIFAR testing data and its corrupted versions termed CIFAR-*-C 
\citep{hendrycks_benchmarking_2018}. To evaluate the model's \gls{OOD} detection performance, we consider two tasks: a  standard \gls{OOD} task using \glsunset{SVHN}\gls{SVHN} as the \gls{OOD} dataset for a model trained on CIFAR-10/-100, and a difficult \gls{OOD} task using CIFAR-100 as the \gls{OOD} dataset for a model trained on CIFAR-10, and vice versa. We compute the uncertainty score for \gls{OOD} using the Dempster-Shafer metric as introduced in \citep{sensoy_evidential_2018}, which empirically leads to better performance for distance-aware models (see Appendix \ref{sec:exp_app}
).
Table \ref{tb:cifar10}-\ref{tb:cifar100} reports the results. As shown, for predictive accuracy, \gls{SNGP} is competitive with that of a deterministic network, and outperforms the other single-model approaches. For calibration error, \gls{SNGP} clearly outperforms the other single-model approaches and is competitive with the deep ensemble. Finally, for \gls{OOD} detection, \gls{SNGP} outperforms not only the deep ensembles and \gls{MCD} approaches that are based on a dense output layer, but also the \gls{MCDGP} and \gls{DUQ} that are based on the \gls{GP} layer, illustrating the importance of the \textit{input distance awareness} property for high-quality performance in uncertainty quantification.
\input{./section/tables/cifar10_app}
\vspace{-.5em}
\input{./section/tables/cifar100_app}

\textbf{Detecting Out-of-Scope Intent in Conversational Language Understanding}
To validate the method beyond image modalities, we also evaluate \gls{SNGP} on a practical language understanding task where uncertainty quantification is of natural importance: dialog intent detection 
\citep{larson_evaluation_2019, vedula_towards_2019, yaghoub-zadeh-fard_user_2020, zheng_out--domain_2020}.
In a goal-oriented dialog system (e.g. chatbot) built for a collection of in-domain services, it is important for the model to understand if an input natural utterance from an user is in-scope (so it can activate one of the in-domain services) or out-of-scope (where the model should abstain). To this end, we consider training an intent understanding model using the CLINC \gls{OOS} intent detection benchmark dataset \citep{larson_evaluation_2019}. Briefly, the \gls{OOS} dataset contains data for 150 in-domain services with 150 training sentences in each domain, and also 1500 natural out-of-domain utterances. 
We train the models only on in-domain data, and evaluate their predictive accuracy on the in-domain test data, their calibration and \gls{OOD} detection performance on the combined in-domain and out-of-domain data. The results are in Table \ref{tb:clinc}. As shown, consistent with the previous vision experiments, \gls{SNGP} is competitive in predictive accuracy when compared to a deterministic baseline, and outperforms other approaches in calibration and \gls{OOD} detection.

\input{./section/tables/clinc_app}

%% file: section/tables/methods.tex
\begin{table}[ht]
    \centering
    \scalebox{0.65}{
    \begin{tabular}{c|cccc}
    \toprule
      & Additional & Output & Ensemble & Multi-pass \\
     Methods & Regularization & Layer & Training & Inference \\
     \midrule
     Deterministic & - & Dense & - & - \\
     \midrule
     \gls{MCD} & Dropout & Dense & - & Yes \\
     Deep Ensemble & - & Dense & Yes & Yes \\
     \midrule
     \gls{MCDGP} & Dropout & GP & - & Yes \\
     \gls{DUQ} & Gradient Penalty & RBF & - & - \\
     \midrule
     {DNN-SN} & Spec Norm & Dense & - & - \\
     {DNN-GP} & - & GP & - & - \\
     \gls{SNGP} & Spec Norm & GP & - & - \\
     \bottomrule
    \end{tabular}
    }
    \captionsetup{justification=centering}
    \caption{Summary of methods used in experiments. 
    Multi-pass Inference refers to whether the method needs to perform multiple forward passes to generate the predictive distribution.}
    \label{tb:methods}
\end{table}

%% file: section/tables/cifar10_app.tex
\begin{table}[ht]
\centering
\resizebox{\textwidth}{!}{  
\begin{tabular}{ccc|cc|cc|cc|c}
\toprule
& \multicolumn{2}{c}{Accuracy ($\uparrow$)} & 
\multicolumn{2}{c}{ECE ($\downarrow$)} &
\multicolumn{2}{c}{NLL ($\downarrow$)} &
\multicolumn{2}{c}{OOD AUPR ($\uparrow$)} &
Latency ($\downarrow$)
\\
Method  & Clean & Corrupted & Clean & Corrupted & Clean & Corrupted & 
SVHN & CIFAR-100 & {\small (ms / example)}  \\
\midrule
Deterministic & 96.0 $\pm$ 0.01 & 72.9 $\pm$ 0.01 & 0.023 $\pm$ 0.002 & 0.153 $\pm$ 0.011 & 0.158 $\pm$ 0.01 & 1.059 $\pm$ 0.02 &
0.781 $\pm$ 0.01 & 0.835 $\pm$ 0.01 & \textbf{3.91}
\\
\midrule
\glsunset{MCD}\gls{MCD} & 96.0 $\pm$ 0.01 & 70.0 $\pm$ 0.02 & 0.021 $\pm$ 0.002 & 0.116 $\pm$ 0.009 & 0.173 $\pm$ 0.01 & 1.152 $\pm$ 0.01 & 0.971 $\pm$ 0.01 & 0.832 $\pm$ 0.01 & 27.10\\
Deep Ensembles & \textbf{96.6 $\pm$ 0.01} & \textbf{77.9 $\pm$ 0.01} & \textbf{0.010 $\pm$ 0.001}  & \textbf{0.087 $\pm$ 0.004}  & \textbf{0.114 $\pm$ 0.01} & \textbf{0.815 $\pm$ 0.01} & 0.964 $\pm$ 0.01 & \underline{0.888 $\pm$ 0.01} & 38.10\\
\midrule
\glsunset{MCDGP}\gls{MCDGP} & 95.5 $\pm$ 0.02 & 70.0 $\pm$ 0.01 & 0.024 $\pm$ 0.004 & 0.100 $\pm$ 0.007 &  0.172 $\pm$ 0.01 & 1.157 $\pm$ 0.01 &
0.960 $\pm$ 0.01 & 0.863 $\pm$ 0.01 & 29.53\\
\glsunset{DUQ}\gls{DUQ}  & 94.7 $\pm$ 0.02 & 71.6 $\pm$ 0.02 & 0.034 $\pm$ 0.002 & 0.183 $\pm$ 0.011 & 0.239 $\pm$ 0.02 & 1.348 $\pm$ 0.01 & 
0.973 $\pm$ 0.01 & 0.854 $\pm$ 0.01 & 8.68\\
\midrule
DNN-SN & 96.0 $\pm$ 0.01 & 72.5 $\pm$ 0.01 & 0.025 $\pm$ 0.004 & 0.178 $\pm$ 0.013 &  0.171 $\pm$ 0.01 & 1.306 $\pm$ 0.01 & 0.974 $\pm$ 0.01 & 0.859 $\pm$ 0.01 & 5.20 \\
DNN-GP & \underline{95.9 $\pm$ 0.01} & 71.7 $\pm$ 0.01 & 0.029 $\pm$ 0.002  & 0.175 $\pm$ 0.008 &  0.221 $\pm$ 0.02 & 1.380 $\pm$ 0.01 & \underline{0.976 $\pm$ 0.01} & 0.887 $\pm$ 0.01 & 5.58 \\
\gls{SNGP} (Ours) & \underline{95.9 $\pm$ 0.01} & \underline{74.6 $\pm$ 0.01} & \underline{0.018 $\pm$ 0.001} & \underline{0.090$\pm$ 0.012} & \underline{0.138 $\pm$ 0.01} & \underline{0.935 $\pm$ 0.01} & \textbf{0.990 $\pm$ 0.01} & \textbf{0.905 $\pm$ 0.01} & 6.25 \\
\bottomrule
\end{tabular}
}
\caption{Results for Wide ResNet-28-10 on CIFAR-10, averaged over 10 seeds.  
}
\label{tb:cifar10}
\end{table}

%% file: section/tables/cifar100_app.tex
\begin{table}[ht]
\centering
\resizebox{\textwidth}{!}{  
\begin{tabular}{ccc|cc|cc|cc|c}
\toprule
& \multicolumn{2}{c}{Accuracy ($\uparrow$)} & 
\multicolumn{2}{c}{ECE ($\downarrow$)} &
\multicolumn{2}{c}{NLL ($\downarrow$)} &
\multicolumn{2}{c}{OOD AUPR ($\uparrow$)} &
Latency ($\downarrow$)
\\
Method  & Clean & Corrupted & Clean & Corrupted & Clean & Corrupted & 
SVHN & CIFAR-10 & {\small (ms / example)}  \\
\midrule
Deterministic & 79.8 $\pm$ 0.02 & \underline{50.5 $\pm$ 0.04} & 0.085 $\pm$ 0.004 & 0.239 $\pm$ 0.020 & 0.872 $\pm$ 0.01 & 2.756 $\pm$ 0.03 & 0.882 $\pm$ 0.01 & 0.745 $\pm$ 0.01 & \textbf{5.20} \\
\midrule
\gls{MCD} & 79.6 $\pm$ 0.02 & 42.6 $\pm$ 0.08 & 0.050 $\pm$ 0.003 & 0.202 $\pm$ 0.010 &  0.825 $\pm$ 0.01 & 2.881 $\pm$ 0.01 & 0.832 $\pm$ 0.01 & 0.757 $\pm$ 0.01 & 
46.79\\
Deep Ensemble & \textbf{80.2 $\pm$ 0.01} & \textbf{54.1 $\pm$ 0.04} & \textbf{0.021 $\pm$ 0.004} & \underline{0.138$\pm$ 0.013} &  \textbf{0.666 $\pm$ 0.02} & \textbf{2.281 $\pm$ 0.03} & \underline{0.888 $\pm$ 0.01} & \underline{0.780 $\pm$ 0.01} & 42.06\\
\midrule
\gls{MCDGP} & 79.5$\pm$ 0.04 & 45.0 $\pm$ 0.05 & 0.085 $\pm$ 0.005 & 0.159 $\pm$ 0.009 &  0.937 $\pm$ 0.01 & \underline{2.584 $\pm$ 0.02} & 0.873 $\pm$ 0.01 & 0.754 $\pm$ 0.01 & 
44.20 \\
\gls{DUQ}  & 78.5 $\pm$ 0.02 & 50.4 $\pm$ 0.02 & 0.119 $\pm$ 0.001 & 0.281 $\pm$ 0.012 &  0.980 $\pm$ 0.02 & 2.841 $\pm$ 0.01 & 0.878 $\pm$ 0.01 & 0.732 $\pm$ 0.01 & 6.51\\
\midrule
DNN-SN & \underline{79.9 $\pm$ 0.02} & 48.6 $\pm$ 0.02 & 0.098$\pm$ 0.004 & 0.272$\pm$ 0.011 &  0.918 $\pm$ 0.01 & 3.013$\pm$ 0.01 & 0.879$\pm$ 0.03 & 0.745$\pm$ 0.01 & 6.20 \\
DNN-GP & 79.2 $\pm$ 0.03 & 47.7 $\pm$ 0.03 & 0.064$\pm$ 0.005 & 0.166$\pm$ 0.003 & 0.885$\pm$ 0.009 & 2.629$\pm$ 0.01 &  0.876$\pm$ 0.01 & 0.746$\pm$ 0.02 & 6.82 \\
\gls{SNGP} (Ours) & \underline{79.9 $\pm$ 0.03} & 49.0 $\pm$ 0.02 & \underline{0.025 $\pm$ 0.012} & \textbf{0.117 $\pm$ 0.014} &  \underline{0.847 $\pm$ 0.01} & 2.626 $\pm$ 0.01 & \textbf{0.923 $\pm$ 0.01} & \textbf{0.801 $\pm$ 0.01} & 6.94\\
\bottomrule
\end{tabular}
}
\caption{Results for Wide ResNet-28-10 on CIFAR-100, averaged over 10 seeds.}
\label{tb:cifar100}
\end{table}

%% file: section/tables/clinc_app.tex
\begin{table}[!h]
\centering
\scalebox{0.65}{
\begin{tabular}{cc|c|c|cc|c}
\toprule
& \multicolumn{1}{c}{Accuracy ($\uparrow$)} & 
\multicolumn{1}{c}{ECE ($\downarrow$)} &
\multicolumn{1}{c}{NLL ($\downarrow$)} &
\multicolumn{2}{c}{OOD} & Latency ($\downarrow$) \\
Method  &   &   &  & 
AUROC ($\uparrow$) & AUPR ($\uparrow$) & 
{\small (ms / example)}\\
\midrule
Deterministic & 96.5 $\pm$ 0.11 & 0.024 $\pm$ 0.002 & 3.559 $\pm$ 0.11 & 0.897 $\pm$ 0.01 & 0.757 $\pm$ 0.02 & \textbf{10.42} \\
\midrule
\gls{MCD} & 96.1 $\pm$ 0.10 & 0.021 $\pm$ 0.001  & 1.658 $\pm$ 0.05 & 0.938 $\pm$ 0.01 & 0.799 $\pm$ 0.01 & 85.62 \\
Deep Ensemble & \textbf{97.5 $\pm$ 0.03} & \textbf{0.013 $\pm$ 0.002} & \textbf{1.062 $\pm$ 0.02} & \underline{0.964 $\pm$ 0.01} & \underline{0.862 $\pm$ 0.01} & 84.46 \\
\midrule
\gls{MCDGP} & 95.9 $\pm$ 0.05 & 0.015 $\pm$ 0.003 & 1.664 $\pm$ 0.04 & 0.906 $\pm$ 0.02 & 0.803 $\pm$ 0.01 & 88.38\\
\gls{DUQ} & 96.0 $\pm$ 0.04 & 0.059 $\pm$ 0.002  & 4.015 $\pm$ 0.08 & 0.917 $\pm$ 0.01 & 0.806 $\pm$ 0.01 & 15.60\\
\midrule
DNN-SN & 95.4 $\pm$ 0.10 & 0.037 $\pm$ 0.004 & 3.565 $\pm$ 0.03 & 0.922 $\pm$ 0.02 & 0.733 $\pm$ 0.01 & 17.36  \\
DNN-GP & 95.9 $\pm$ 0.07 & 0.075 $\pm$ 0.003 & 3.594 $\pm$ 0.02 & 0.941 $\pm$ 0.01 & 0.831 $\pm$ 0.01 & 18.93  \\
\gls{SNGP} & \underline{96.6 $\pm$ 0.05} & \underline{0.014 $\pm$ 0.005} & \underline{1.218 $\pm$ 0.03} & \textbf{0.969 $\pm$ 0.01} & \textbf{0.880 $\pm$ 0.01} & 17.36 \\
\bottomrule
\end{tabular}
}
\caption{\small{
Results for BERT$_{\texttt{Base}}$ on CLINC \gls{OOS}, averaged over 10 seeds.
}}
\vspace{-1em}
\label{tb:clinc}
\end{table}

%% file: section/conclusion.tex
We propose SNGP, a simple approach to improve a single deterministic \gls{DNN}'s ability in predictive uncertainty estimation. It makes minimal changes to the architecture and training/prediction pipeline of a deterministic \gls{DNN}, only adding spectral normalization to the hidden mapping, and replacing the dense output layer with a random feature layer that approximates a \gls{GP}. We theoretically motivate \textit{input distance awareness}, the key design principle behind \gls{SNGP}, via a learning-theoretic analysis of the uncertainty estimation problem. We also 
propose a closed-form approximation method to make the GP posterior end-to-end trainable in linear time with the rest of the neural network. On a suite of vision and language understanding tasks and on modern architectures (ResNet and BERT), 
\gls{SNGP} is competitive with a deep ensemble in prediction, calibration and out-of-domain detection, and outperforms other single-model approaches. 

A central observation we made in this work is that \textit{good  representational learning is important for good uncertainty quantification}. In particular, we highlighted  \textit{bi-Lipschitz} (Equation (\ref{eq:dp})) as an important condition for the learned representation of a \gls{DNN} to attain high-quality uncertainty performance, and proposed spectral normalization as a simple approach to ensure such property in practice. However, it is worth noting that there exists other representation learning techniques, e.g., data augmentation or unsupervised pretraining, that are known to also improve a network's uncertainty performance \citep{hendrycks_using_2019, hendrycks_augmix_2020}. Analyzing whether and how these approaches contribute to improve a \gls{DNN} \textit{bi-Lipschitz} condition, and whether the \textit{bi-Lipschitz} condition is sufficient in explaining these methods' success, are interesting avenues of future work. Furthermore, we note that the spectral norm bound $\alpha < 1$ in Proposition \ref{thm:resnet_lipschitz} forms only a sufficient condition for ensuring bi-Lipschitz \citep{behrmann_invertible_2019}. In practice, we observed that for convolutional layers, a looser norm bound is needed for state-of-the-art performance (see Section \ref{sec:exp_app}), raising questions of whether the current regularization approach is precise enough in controlling the spectral norm of a convolutional kernel, or if there is an alternative mechanism at play in ensuring the bi-Lipschitz criterion. Finally, from a probabilistic learning perspective, SNGP focuses on learning a single high-quality model $p_\theta(y|\bx)$ for a deterministic representation.  Therefore we expect it to provide complementary benefits to approaches such as (efficient) ensembles and Bayesian neural networks \citep{dusenberry_efficient_2020, lakshminarayanan_simple_2017, wen_batchensemble_2020} which marginalize over the representation parameters as well. 

%% file: section/acknowledge.tex
\textbf{Acknowledgements}
We 
would like to thank Kevin Murphy, Deepak Ramachandran, Jasper Snoek, and Timothy Nguyen at Google Research for the insightful comments and fruitful discussion. 

%% file: section/impact.tex
This work proposed a simple and practical methodology to improve the uncertainty estimation performance of a deterministic deep learning model. Experiment results showcased the method's ability in improving model performance in calibration and \gls{OOD} detection while maintaining similar level of accuracy and latency, therefore illustrating its feasibility for industrial-scale applications. We hope the proposed approach can be used to bring concrete improvements to AI-driven, socially-relevant services where uncertainty is of natural importance. Examples include medical and policy decision making, online toxic comment management, fairness-aware recommendation systems, etc.



Nonetheless, we do not claim that the improvement illustrated in this paper
solve the problem of model uncertainty entirely. This is because
the analysis and experiments in this study may not capture the full complexity of the 
real-world use cases, 
and there will always be room for improvement. 
Designers of machine learning systems are encouraged to proactively confront the shortcomings of model uncertainty and the underlying models that generate these confidences. Even with a proper user interface, there is always room to misinterpret model outputs and probabilities, such as with nuanced applications such as election predictions, and users of these models should to be properly trained to take these factors into account.

%% file: section/figures/2d_exp_oval_app.tex
\begin{figure}[th!]
    \centering
    \subcaptionbox{Gaussian Process\label{fig:2d_exp_oval_gp_app}}{
    \includegraphics[width=0.25\columnwidth]{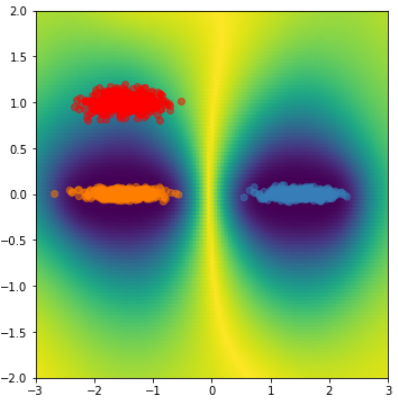}}
    \subcaptionbox{Deep Ensemble\label{fig:2d_exp_oval_deepens_app}}{
    \includegraphics[width=0.25\columnwidth]{./figures/2d/oval_deep_ens}}
    \subcaptionbox{MC Dropout\label{fig:2d_exp_oval_mc_app}}{
    \includegraphics[width=0.25\columnwidth]{./figures/2d/oval_mc}}
    
    \subcaptionbox{DNN-SN \label{fig:2d_exp_oval_deepsn_app}}{
    \includegraphics[width=0.25\columnwidth]{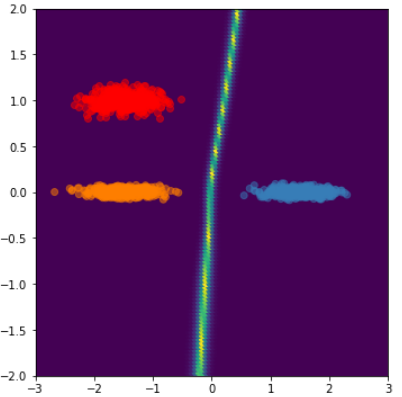}}
    \subcaptionbox{DNN-GP \label{fig:2d_exp_oval_deepgp_app}}{
    \includegraphics[width=0.25\columnwidth]{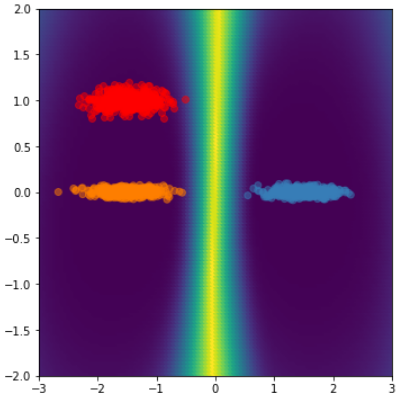}}
    \subcaptionbox{\gls{SNGP} (Ours)\label{fig:2d_exp_oval_sngp_app}}{
    \includegraphics[width=0.25\columnwidth]{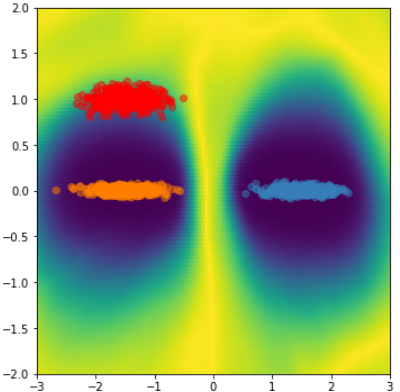}}
   \vspace{-0.5em}
    \caption{
    \small{
    The uncertainty surface of a \gls{GP} and different DNN approaches on the \textit{two ovals} 2D classification benchmarks. The uncertainty is computed in terms of the distance of the maximum predictive probability from 0.5, i.e. $u(\bx) = 1 - 2 * |p(\bx) - 0.5|$. Background color represents the  estimated model uncertainty (See \ref{fig:2d_exp_oval_sngp} for color map). 
    }}
     \vspace{-0.5em}
    \label{fig:2d_exp_oval_app}
\end{figure}

%% file: section/figures/2d_exp_moon_app.tex
\begin{figure}[th!]
    \centering
    \subcaptionbox{Gaussian Process\label{fig:2d_exp_moon_gp_app}}{
    \includegraphics[width=0.25\columnwidth]{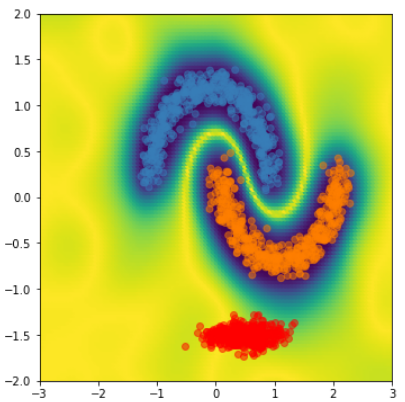}}
    \subcaptionbox{Deep Ensemble\label{fig:2d_exp_moon_deepens_app}}{
    \includegraphics[width=0.25\columnwidth]{./figures/2d/moon_deep_ens}}
    \subcaptionbox{MC Dropout\label{fig:2d_exp_moon_mc_app}}{
    \includegraphics[width=0.25\columnwidth]{./figures/2d/moon_mc}}
    
    \subcaptionbox{DNN-SN \label{fig:2d_exp_moon_deepsn_app}}{
    \includegraphics[width=0.25\columnwidth]{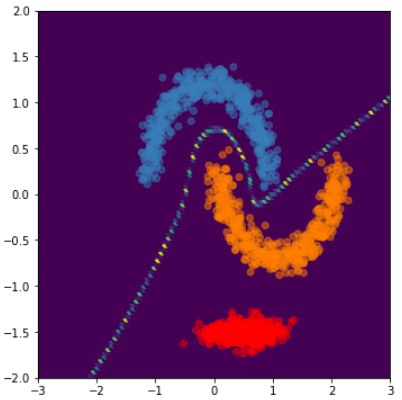}}
    \subcaptionbox{DNN-GP \label{fig:2d_exp_moon_deepgp_app}}{
    \includegraphics[width=0.25\columnwidth]{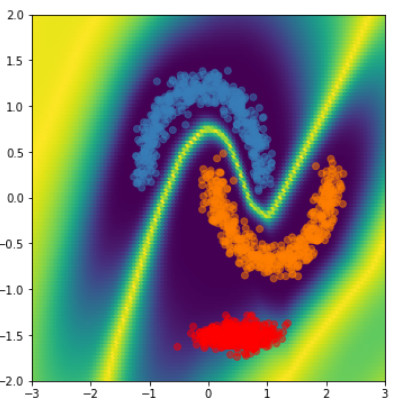}}
    \subcaptionbox{\gls{SNGP} (Ours)\label{fig:2d_exp_moon_sngp_app}}{
    \includegraphics[width=0.25\columnwidth]{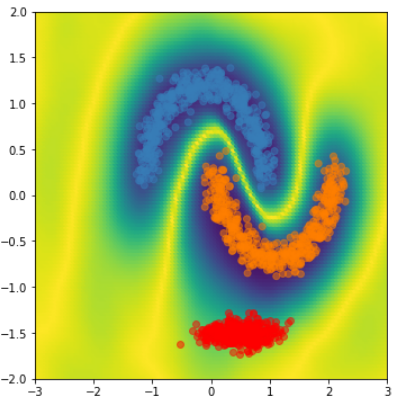}}
   \vspace{-0.5em}
    \caption{
    \small{
    The uncertainty surface of a \gls{GP} and different DNN approaches on the \textit{two moons} 2D classification benchmarks. The uncertainty is computed in terms of the distance of the maximum predictive probability from 0.5, i.e. $u(\bx) = 1 - 2 * |p(\bx) - 0.5|$. Background color represents the  estimated model uncertainty (See Figures \ref{fig:2d_exp_moon_sngp} and \ref{fig:2d_exp_oval_sngp} for color map).
    }}
     \vspace{-0.5em}
    \label{fig:2d_exp_moon_app}
\end{figure}

%% file: section/proof/minimax_lemma.tex
This proof is an application of the generalized maximum entropy theorem to the case of Bregman score. We shall first state the generalized maximum entropy theorem to make sure the proof is self-contained. Briefly, the generalized maximum entropy theorem verifies that for a general scoring function $s(p, p^*|\bx)$ with entropy function $H(p|\bx)$, the maximum-entropy distribution $p' = \underset{p}{argsup} \, H(p|\bx)$ attains the minimax optimality :

\begin{theorem}[Maximum Entropy Theorem for General Loss  \citep{grunwald_game_2004}]
Let $\Psc$ be a convex, weakly closed and tight set of distributions. Consider a general score function $s(p, p^*|\bx)$ with an associated entropy function defined as $H(p|\bx) = \inf_{p^*\in \Psc^*} s(p, p^*|\bx)$. Assume below conditions on $H(p|\bx)$ hold:
\begin{itemize}
    \item (Well-defined) For any $p \in \Psc$, $H(p|\bx)$ exists and is finite. 
    \item (Lower-semicontinous) For a weakly converging sequence $p_n \rightarrow p_0 \in \Psc$ where $H(p_n|\bx)$ is bounded below, we have $s(p, p_0|\bx) \leq \liminf_{n\rightarrow \infty} s(p, p_n|\bx)$ for all $p \in \Psc$.
\end{itemize}
Then there exists an maximum-entropy distribution $p'$ such that 
$$p' = \sup_{p \in \Psc}H(p)=\sup_{p \in \Psc} \inf_{p^*\in \Psc^*} s(p, p^*|\bx) = \inf_{p \in \Psc} \sup_{p^* \in \Psc^*} s(p, p^*|\bx).$$
\label{thm:minimax_general}
\end{theorem}
Above theorem states that the maximum-entropy distribution attains the minimax optimality for a scoring function $s(p, p^*|\bx)$, assuming its entropy function satisfying certain regularity conditions. Authors of \citep{grunwald_game_2004}  showed that the entropy function of a Bregman score satisfies conditions in Theorem 1. Consequently, to show that the discrete uniform distribution is minimax optimal for Bregman score at $\bx \not\in \Xsc_{\texttt{IND}}$, we only need to show discrete uniform distribution is the maximum-entropy distribution.

Recall the definition of the \textit{strictly} proper Bregman score \citep{parry_proper_2012}:
\begin{align}
    s(p, p^*|\bx)= \sum_{k=1}^K 
    \Big\{
    [p^*(y_k|\bx) - p(y_k|\bx)]\psi'(p^*(y_k|\bx)) - \psi(p^*(y_k|\bx))
    \Big\}
\end{align}
where $\psi$ is differentiable and \textit{strictly} concave. Moreover, its entropy function is:
\begin{align}
    H(p|\bx) = -\sum_{k=1}^K \psi(p(y_k|\bx))
    \label{eq:bregman_entropy}
\end{align}
Our interest is to show that for $\bx \in \Xsc_{\texttt{OOD}}$, the maximum-entropy distribution for the Bregman score is the discrete uniform distribution $p(y_k|\bx)=\frac{1}{K}$. To this end, we notice that in the absence of any information, the only constraint on the predictive distribution is that $\sum_k p(y_k|\bx) = 1$. Therefore, denoting $p(y_k|\bx)=p_k$, we can set up the optimization problem with respect to Bregman entropy (\ref{eq:bregman_entropy}) using the Langrangian form below:
\begin{align}
L(p|\bx) &= H(p|\bx) + \lambda * (\sum_k p_k - 1)
= -\sum_{k=1}^K \psi(p_k) + \lambda * (\sum_k p_k - 1)
\end{align}
Taking derivative with respect to $p_k$ and $\lambda$:
\begin{align}
    \deriv{p_k} L &= -\psi'(p_k) + \lambda = 0\\
    \deriv{\lambda} L &= \sum_{k=1}^K p_k - 1 = 0
\end{align}
Notice that since $\psi(p)$ is \textit{strictly} concave, the function $\psi'(p)$ is monotonically decreasing and therefore invertible.
As a result, to solve the maximum entropy problem, we can solve the above systems of equation by finding a inverse function $\psi^{' -1}(p)$, which lead to the simplification:
\begin{align}
    p_k &= \psi^{' -1}(\lambda);  \quad \quad \sum_{k=1}^K p_k = 1.
\end{align}
Above expression essentially states that all $p_k$'s should be equal and sum to 1. The only distribution satisfying the above is the discrete uniform distribution, i.e.,  $p_k = \frac{1}{K} \; \forall k$.